\documentclass[journal]{IEEEtran}

\usepackage{cite}
\usepackage{amsmath,amssymb,amsfonts}
\usepackage{graphicx}
\usepackage{textcomp}
\usepackage{xcolor}
\usepackage{booktabs}
\usepackage{multirow}
\usepackage{array}
\usepackage{url}
\usepackage{tikz}
\usepackage{pgfplots}
\pgfplotsset{compat=1.17}
\usetikzlibrary{arrows.meta,positioning,fit,backgrounds,calc}

\tikzset{
  block/.style={rectangle, rounded corners=2pt, draw=black!70, fill=#1,
                minimum height=7mm, inner sep=4pt, align=center, font=\scriptsize},
  stage/.style={block=blue!8},
  qr/.style={block=orange!14},
  cls/.style={block=teal!12},
  io/.style={rectangle, draw=black!55, fill=black!4, minimum height=7mm,
             inner sep=4pt, align=center, font=\scriptsize},
  ar/.style={-{Stealth[length=2mm]}, draw=black!70, line width=0.5pt},
  grp/.style={draw=black!30, dashed, rounded corners=3pt, inner sep=4pt},
}

\newcommand{\TQRNNd}{TQRNN$_{30\mathrm{d}}$}
\newcommand{\TQRNNh}{TQRNN$_{70\mathrm{h}}$}

\def\BibTeX{{\rm B\kern-.05em{\sc i\kern-.025em b}\kern-.08em
    T\kern-.1667em\lower.7ex\hbox{E}\kern-.125emX}}

\begin{document}

\title{Long-Horizon Transformer–Quantile Fault
Prediction for Multi-Site Industrial Predictive Maintenance.}

\author{David~J.~Poland,~Daniele~Ravi,~and~Na~Helian%
\thanks{The authors are with the Department of Computer Science,
University of Hertfordshire, Hatfield AL10~9AB, U.K.
(e-mail: d.j.poland@herts.ac.uk; d.ravi@herts.ac.uk; n.helian@herts.ac.uk).}}

\markboth{}{Poland \MakeLowercase{\textit{et al.}}: Long-Horizon Transformer--Quantile Fault Prediction}

\maketitle

\begin{abstract}
Long-horizon predictive maintenance requires a model to distinguish slowly evolving
degradation from normal operating-regime variation over planning windows measured in
days rather than hours. This paper evaluates whether an explicit conditional-quantile
representation provides an informative classifier interface for that problem. The
proposed \TQRNNd{} framework combines a dual-stage quantile-regression neural network
(QRNN) feature extractor with a multi-stream temporal-fusion classifier. Each hourly
word of 81-channel machine behaviour is mapped to a 324-dimensional quantile-state
representation; 720 ordered hourly words form the 30-day document supplied to the
long-horizon model. The classifier fuses quantile states with dynamic covariates,
channel-level static metadata, and a 168-hour latent-history stream using gated
residual processing, causal recurrent encoding, and metadata-conditioned cross-modal
attention. A bounded instability-aware signal derived from sustained one-word-ahead
prediction-error divergence provides auxiliary memory modulation at the longest
horizon. Evaluation uses a machine-disjoint 43/14/15 train/validation/test allocation
across 72 machines in nine manufacturing facilities. At 30 days,
\TQRNNd{} achieves 79.97\% F1, 80.18\% recall, 81.82\% precision, 82.39\% accuracy,
and 0.820 ROC--AUC. It leads all 18 evaluated baselines at the 7-, 14-, and 30-day
fixed-threshold comparisons; the F1 advantage over the strongest baseline is retained
at all three horizons and is largest at 14 days. The evidence supports held-out
machine performance within the observed homogeneous nine-facility fleet, but does not
establish unseen-site, cross-equipment, or cross-sector generalisation.
\end{abstract}


\section{Introduction}
\IEEEPARstart{H}{ighly} automated manufacturing systems depend on asset
availability, process stability, and timely maintenance intervention. Corrective
maintenance reacts after failure, while preventive maintenance acts on fixed
intervals that may not reflect the actual condition of a machine
\cite{Sharma,Sellitto}. Predictive maintenance (PdM) seeks to close this gap by
estimating equipment health from sensor evidence before failure occurs
\cite{Chen,Abidi}. Recent advances in embedded sensing, edge acquisition, and
industrial connectivity have made high-volume multivariate monitoring routine
\cite{Yazdi}. However, translating this data volume into reliable
long-horizon maintenance decisions remains challenging.

\subsection*{Why long horizons, and why multiple sites}
Most PdM systems reported in the literature are either short-horizon, evaluated
on benchmark datasets, or designed around individual assets rather than
multi-site industrial deployments \cite{Qiu,Islam,Serradilla,Fernandes}.
This distinction is operationally significant. A \emph{satellite facility}
network---several geographically separated plants operated by one organisation
and producing the same product family---is a common structure in high-volume
discrete manufacturing, and it changes what a maintenance forecast is
\emph{for}. In a single-site setting, a fault warning triggers a local repair
decision. In a satellite network, the same warning is an input to a fleet-level
scheduling problem: which plant absorbs demand while another is down, when
shared spare-part inventory should be repositioned, and how maintenance windows
are sequenced so that aggregate output is preserved. These decisions are taken
on procurement and production-planning timescales of weeks, which is precisely
why a 30-day horizon---rather than an hour-ahead alarm---is the operationally
binding quantity here. Reviews of industrial PdM adoption consistently identify
this gap between asset-level detection and fleet-level planning as a barrier to
realizing the promised benefits \cite{Pozzi,Meddaoui,Islam}.

Three difficulties intensify as the horizon extends from hours to weeks, and each
is documented in the prior literature rather than assumed here. First, incipient
degradation is expressed as gradual distributional change---variance growth,
tail movement, and envelope widening---rather than as large point anomalies, so
detectors tuned to threshold breaches lose sensitivity at long lead times
\cite{Wu,Carrasco,Ong}. Second, heterogeneous sensor groups sampled at
incompatible rates must be fused causally, and naive resampling or
window-overlapping evaluation is a well-documented source of optimistic bias in
industrial time-series studies \cite{Carrasco,Oliosi}. Third, long windows
increase exposure to regime transitions driven by wear progression, lubrication
change, thermal cycling, tooling drift, and material changeover, so the mapping
from sensor state to fault probability is itself non-stationary
\cite{Mitici,Ong}. Recurrent and attention-based classifiers model
sequential structure well but do not, by construction, preserve distributional
or uncertainty information, nor separate sustained degradation-like divergence
from isolated transients.

\subsection*{Long-Horizon Context and Research Gap}
Long-horizon prediction has an established literature, and the contribution of
this paper is best understood against it rather than against short-horizon work
alone. Multi-horizon forecasting architectures such as the Temporal Fusion
Transformer \cite{Lim2021} and long-range spatiotemporal transformers
\cite{Grigsby} explicitly target extended horizons, and probabilistic RUL
prognostics \cite{Mitici} and multi-scale degradation models \cite{Fernandes}
forecast weeks or cycles ahead. Three limitations, however, recur. (i) Most
long-horizon work targets \emph{regression} onto a degradation index or RUL,
which presupposes a monotone health indicator and run-to-failure trajectories;
in a live production fleet, assets are repaired before failure and such
trajectories are largely unavailable. (ii) Long-horizon evaluations are
frequently conducted on benchmark corpora with a single asset type and a fixed
operating regime, so robustness to cross-site regime variation is untested
\cite{Qiu,Serradilla}. (iii) Where uncertainty is modelled, it is usually
reported as an \emph{output} interval rather than exploited as an
\emph{input representation} for a downstream decision. The present work
addresses a task of the third kind: it retains a distributional description of
the input, but targets a decision-relevant binary event within a planning
window, under multi-site regime variation. These studies motivate comparison with long-horizon and uncertainty-aware
baselines, but their numerical results are not treated as directly comparable
cross-study evidence because the datasets, targets, and validation protocols
differ.

\subsection*{Scope of the task}
This paper presents the Transformer Quantile Regression Neural Network for 30-day
horizons (\TQRNNd{}), a long-horizon framework for industrial fault-occurrence
forecasting. The primary evaluated task is supervised Normal/Abnormal forecasting:
given a completed 30-day observation document ending at reference time $t_0$,
predict whether at least one confirmed fault event occurs within a future horizon
$H\in\{7,14,30\}$ days. Internally, the classifier preserves hourly forecast-
interval structure over the selected horizon; the headline classification metrics
reported here are horizon-level decisions derived from that structured output.
The task is therefore distinct from direct RUL regression or degradation-index
regression.

The model converts multi-rate machine data into hourly word and document
representations (Section~\ref{sec:data}) and extends the earlier TQRNN and
\TQRNNh{} stages \cite{poland} by increasing the long-horizon sensing interface to
81 physical channels and using a multi-stream Temporal Fusion Transformer (TFT)
classifier with an auxiliary instability-aware memory-modulation pathway. Throughout,
QRNN denotes a quantile-regression neural network, MLP a multilayer perceptron, GRN
a gated residual network, GRU a gated recurrent unit, and PLC a programmable logic
controller.

\noindent\textbf{Horizon definition.}
The underlying research programme progresses from short-horizon TQRNN prediction,
through the intermediate \TQRNNh{} regime, to the long-horizon \TQRNNd{} regime.
This article focuses on the completed day-scale evaluation at 7, 14, and 30 days;
\TQRNNh{} is retained only as an architectural reference in the ablation analysis.
No 1-day headline result is claimed in this manuscript.

\noindent\textbf{Reported horizons.}
The cross-model comparison in Table~\ref{tab:comparison_multi}, the day-scale
pipeline ablation in Table~\ref{tab:quantile_sens}, and the principal narrative
all report the same 7-, 14-, and 30-day horizons.

\noindent\textbf{Contributions.}
We distinguish the established TFT and quantile-regression components from the
specific way they are coupled in this work. The contributions are:
\begin{enumerate}
    \item \textbf{Conditional quantiles as a long-horizon classifier interface.}
    The classifier receives a learned conditional-quantile state rather than a raw
    or point-summarised sensor window. Across the completed 7-, 14-, and 30-day
    comparison, the advantage over the strongest baseline is retained at every
    evaluated horizon, although it is not monotonic and is largest at 14 days.
    \item \textbf{A dual-stage quantile representation for the 81-channel
    long-horizon regime.} A broad ten-quantile first stage is followed by a
    four-quantile mid-tail refinement stage, yielding the 324-dimensional
    word-level representation supplied to the classifier.
    \item \textbf{An auxiliary instability-aware memory-modulation pathway.}
    Sustained local one-word-ahead prediction-error divergence is converted to a
    bounded modulation signal used to support temporal memory retention. The
    pathway is explicitly heuristic: it is not a Lyapunov estimator or a separate
    fault classifier, and its independent performance contribution was not isolated
    in a dedicated with/without ablation in the completed experimental programme.
    \item \textbf{A machine-disjoint multi-site evaluation at deployment scale.}
    The final long-horizon regime uses 72 machines across nine facilities, with
    43/14/15 machines assigned exclusively to training, validation, and testing.
    This prevents raw time-series measurements from the same machine appearing in
    more than one partition while preserving representation from the observed
    facility population in each subset.
\end{enumerate}

\noindent We do not claim unseen-site, cross-sector, or cross-equipment-type
transfer. Section~\ref{sec:limits} states the boundaries supported by the
experimental design.

The remainder of the paper is organised as follows. Section~\ref{sec:related}
summarises related work. Section~\ref{sec:data} describes the industrial dataset
and temporal representation. Section~\ref{sec:method} presents the proposed
framework. Section~\ref{sec:setup} defines the experimental setup.
Section~\ref{sec:results} reports the results, and Section~\ref{sec:conclusion}
concludes the paper.

\section{Related Work}
\label{sec:related}

\subsection{Data-Driven Predictive Maintenance}
Data-driven PdM has moved from rule-based condition monitoring toward
statistical, machine-learning, and deep-learning methods that infer degradation
from historical and real-time sensor streams \cite{Qiu,Serradilla}, as part
of the broader Industry~4.0 transition \cite{Pozzi} and supported by systematic
reviews of fault-diagnosis and prognosis techniques \cite{Fernandes}.
Recurrent and long short-term memory models are common because they model
sequential dependencies in degradation data \cite{Meddaoui,Mitici}. However, many
applications remain asset-specific or short-horizon, and reported gains on public
benchmarks do not always translate to industrial deployments where machines,
sites, materials, and operating regimes vary. Hybrid architectures such as
CNN--LSTM models have likewise been applied to joint production and
imperfect-maintenance planning \cite{Shoorkand}, but they are evaluated at a
single short operating horizon and offer no analysis of how predictive
performance behaves as the forecast horizon is extended.

Read together, this literature shows three trends and a corresponding gap.
The first trend is architectural: the field has moved from hand-crafted
condition indicators, through recurrent encoders \cite{Meddaoui,Mitici}, to
attention-based and hybrid multi-branch models \cite{Gawde,Shoorkand}, with each
step buying longer effective memory at the cost of more parameters and more
training data. The second is representational: increasing effort is spent on
\emph{what} is fed to the model---multi-sensor fusion, multi-scale decomposition,
and learned embeddings---rather than only on the classifier itself
\cite{Gawde,Fernandes}. The third is evaluative: surveys repeatedly note that
reported accuracies are difficult to compare because datasets, horizons, and
partitioning protocols differ, and that leakage-prone evaluation is common
\cite{Serradilla,Carrasco,Islam}. The gap this leaves is a method that treats the
\emph{input representation} as the primary object of design for long horizons
specifically, and that is validated under a partitioning protocol strict enough
for the claim to survive scrutiny. That is the gap addressed here.

\subsection{Quantile Regression for Prognostics}
Point forecasts can hide distributional widening, asymmetric tails, and
early-stage degradation envelopes. Quantile regression estimates conditional
quantiles rather than a single expected value and is therefore useful when the
shape of the predictive distribution matters \cite{Ong,Mohammad}. In PdM, this
is important because abnormal behaviour may first appear as tail movement or
uncertainty growth before a clear threshold breach. The present work uses
quantile regression not as a final forecaster, but as a feature interface: a
dual-stage QRNN transforms raw multi-sensor behaviour into a compact
distribution-aware state representation for downstream temporal classification.
By encoding how each channel's conditional quantiles evolve, the extractor
captures the \emph{trajectory pattern} of a sensor rather than the raw values
within a single window. This suppresses the sample-level noise of closely spaced
readings and exposes the slow distributional drift associated with incipient
degradation. Section~\ref{sec:whyquantiles} makes the corresponding argument
against the natural alternatives---variance, skewness, and density
features---explicit, since the choice of a quantile encoding over these
requires justification rather than assertion.

\subsection{Long-Horizon Degradation Modelling and RUL Prediction}
\label{sec:rw_longhorizon}
A parallel line of work forecasts degradation over extended horizons, most often
as an RUL regression. Because that literature defines the state of the art
against which a long-horizon claim must be judged, we summarise it and state
what distinguishes the present task.

Transformer-based degradation models address the data-availability problem
directly. ~\cite{ZSTT} propose ZSTT, a zero-shot transformer
that combines approximate Bayesian inference with a Gumbel-based dynamic masking
module so that multisource degradation prediction can be performed without
pretraining on the target system, validated on hydraulic-actuator degradation.
This addresses a constraint the present work does not face---we have labeled
target-system data---but it shares the underlying premise that the input
representation, not merely the sequence model, governs transferability.

Multi-scale feature fusion is the second recurring strategy. Adaptive
dual-branch multi-scale fusion for complex-equipment RUL \cite{DualBranch}
and related parallel multi-scale designs decompose the signal across temporal
resolutions before fusion, on the argument that degradation evidence is
distributed across scales. Our dual-stage quantile extractor is motivated by the
same concern but decomposes across the \emph{distributional} axis at a fixed
temporal resolution rather than across temporal scales.

Cross-condition generalisation is the third. Adversarial domain-adaptation
methods with stage division for ship electric-propulsion RUL \cite{ShipRUL}
align intra- and inter-domain feature distributions so that a model trained
under one operating condition transfers to another. This is a close analogue to the domain-shift problem encountered in a multi-site
fleet. The present work does not perform explicit domain alignment and does not
claim unseen-site transfer: the long-horizon split is machine-disjoint but remains
stratified across the observed facility population.

Finally, multi-sensor selection and graph-structured fusion---as in
ParallelGraphNet-driven sensor optimisation  for high-speed diagnosis
\cite{ParallelGraphNet}---show that channel selection materially affects
diagnostic accuracy in multi-sensor industrial settings. We fix the channel set
by engineering judgement rather than optimising it, which we note as a
limitation in Section~\ref{sec:limits}.

\subsection{Temporal Fusion and Attention Models}
Transformer models provide strong long-range sequence modelling through
attention \cite{Vaswani,Grigsby}. The Temporal Fusion Transformer (TFT)
combines gated residual networks, static covariate conditioning, temporal
variable selection, and interpretable attention for multi-horizon forecasting
\cite{Lim2021}. These properties are attractive for industrial PdM because
machine behaviour depends on both temporal sensor evolution and contextual
factors such as machine identity, facility conditions, material transitions,
and known-ahead production schedules. \TQRNNd{} adapts the TFT idea to a supervised
industrial fault-prediction task, where the primary classifier input is a learned document-level
quantile-state sequence rather than a direct raw sensor sequence.

\subsection{Anomaly Detection Versus Fault Forecasting}
\label{sec:rw_anomaly}
Anomaly detection and fault forecasting are related but distinct, and it is
worth stating plainly which of the two this paper performs. \emph{Anomaly
detection} asks whether the current observation is inconsistent with normal
behaviour: it is a statement about the present, is frequently unsupervised, and
is evaluated on how well it flags the deviation once it is present
\cite{Wu,Oliosi,Carrasco}. \emph{Fault forecasting}, the task addressed here,
asks whether a labelled abnormal event will occur within a future window: it is
a statement about the future, is supervised, and is evaluated on lead time as
well as separability.

\textbf{The distinction in this paper is between anomaly detection and prospective
anomaly prediction, rather than between unrelated tasks.} Conventional anomaly
detection determines whether the current or immediately observed machine state
deviates from learned normal behaviour. In contrast, \TQRNNd{} uses the evolving
distributional structure of the observation document to predict whether abnormal
or fault-associated behaviour will occur within a future forecasting horizon.
The framework therefore moves the anomaly-detection problem forward in time:
the objective is not simply to identify an anomaly once it is observable, but to
recognise precursor changes in multivariate sensor behaviour sufficiently early
to support maintenance intervention before the abnormal event occurs.

This prospective formulation is central to the proposed method. The dual-stage
QRNN representation captures changes in the conditional sensor distributions,
including displacement, widening, and asymmetry across retained quantile states,
while the temporal-fusion classifier determines whether those changes are
consistent with the progression towards a confirmed future abnormal event.
Accordingly, anomaly-related information remains fundamental to the model, but
it is incorporated as predictive evidence rather than as a separate
reconstruction-error, density-threshold, or unsupervised alarm stage at
inference.

Anomaly-detection methods are additionally retained as comparative baselines.
The autoencoder and generative-adversarial-network (GAN) models in
Table~\ref{tab:comparison_multi} are adapted to the same future
fault-occurrence target as the other classifiers, providing a like-for-like test
of whether conventional anomaly-oriented representations remain competitive
when the task is shifted from present-state detection to long-horizon
prediction. The wider anomaly-detection literature also provides an important
methodological foundation for this formulation because progressive degradation
may first become observable as a change in the distribution or temporal
structure of sensor behaviour before a conventional operating threshold is
breached \cite{Wu,Carrasco}. The contribution of \TQRNNd{} is therefore not to
exclude anomaly detection, but to extend its underlying principle towards
causal, long-horizon prediction of future abnormal machine behaviour.

\subsection{Instability Indicators}
At longer horizons, degradation may involve locally unstable behaviour in which
prediction errors diverge over several successive steps rather than appearing as
a single spike. Classical non-linear diagnostics such as Lyapunov exponents
characterise trajectory divergence \cite{Wolf}, but they require phase-space
reconstruction, long stationary records, and careful embedding-parameter
selection, none of which is available in a causal industrial deployment at
scale. We therefore do \emph{not} estimate any dynamical invariant. The gate
described in Section~\ref{sec:gate} is a heuristic detector of sustained
prediction-error growth, and we refer to it throughout as
\emph{instability-aware} rather than chaos-aware, to avoid implying a formal
dynamical-systems result that this paper does not establish.

\section{Industrial Dataset and Temporal Representation}
\label{sec:data}

\subsection{Deployment Context and Asset Description}
\label{sec:context}
The data derive from proprietary industrial sensor installations deployed across nine manufacturing facilities operated by a single organisation. Facility identifiers, geographic identifiers, equipment-specific descriptors, and product-specific parameters were anonymised prior to analysis in accordance with the industrial data-governance agreement. The monitored assets form a homogeneous fleet of industrial production equipment operating within continuous manufacturing environments. Each asset performs repetitive, high-duty-cycle operations under sustained cyclic loading and variable operating conditions.

The full deployment comprises 72 machines distributed across facilities in Europe,
the Middle East, and the United States. Failure-event records and associated sensor
streams span a 24-month operational period from January 2024 to December 2025.
This corresponds to approximately 1,263,168 machine-hours of potential observation
across the complete fleet before preprocessing exclusions. The deployment belongs
to one organisation and one machine/product family; the multi-facility scale
therefore exposes the models to operational variability without constituting a
cross-sector or cross-equipment evaluation.

\subsubsection{Sensor deployment and acquisition architecture}
Sensors follow a fault-informed instrumentation strategy. In the 81-channel
long-horizon configuration, the monitored subsystems include main-motor and cradle
bearing behaviour (Group~1), hydraulic circuits (Groups~2 and~4), thermal channels
(Group~3), tooling contact and actuation behaviour (Group~5), and high-frequency
accelerometer monitoring of drive-train and tooling assemblies (Group~6).

The acquisition network is hierarchical. Machine-local data-acquisition units
(DAUs) collect field signals and forward them to facility edge servers over the
operational-technology network. Buffered and time-stamped data are then transferred
to the central storage and analytics environment through the secured enterprise
connection. The DAUs are synchronised to a common time reference using Precision
Time Protocol (PTP/IEEE~1588), providing timing accuracy sufficient for the
20\,ms analytical reference grid used below.

\subsubsection{Operational sensor-ablation context}
The deployment-wide sensor-ablation analysis provides additional evidence about
instrumentation redundancy, but it is distinct from the 15-machine held-out cohort
used for the formal long-horizon classifier comparison. On the deployment-count
evidence, non-thermal channels retained detection rates of approximately 94\% or
higher under single-sensor removal, whereas the thermal group was less central to
the mechanically coupled fault-discriminative pathway. This supports the use of
cross-group fusion and indicates resilience to isolated sensor loss; it should not
be interpreted as an additional unseen-site generalisation test.

\subsection{Sensor Configuration}
Each machine is instrumented with heterogeneous 12-bit sensors producing integer
readings in the range $[0,4095]$. The long-horizon \TQRNNd{} configuration uses
81 channels across six sensor groups. Table~\ref{tab:sensor_groups} summarises
the sensor composition and native sampling rates.

\begin{table}[t]
\centering
\caption{Sensor group composition and native sampling rates.}
\label{tab:sensor_groups}
\small
\setlength{\tabcolsep}{4pt}
\begin{tabular}{@{}c p{2.8cm} c c c@{}}
\toprule
\textbf{Grp.} & \textbf{Modality} & \textbf{Rate} & \textbf{Period} & \textbf{Channels} \\
\midrule
1 & Vibration / motion        & 1 Hz     & 1 s   & 6  \\
2 & Fluid / flow              & 1 Hz     & 1 s   & 7  \\
3 & Thermal                   & 0.033 Hz & 30 s  & 4  \\
4 & Pressure                  & 50 Hz    & 20 ms & 18 \\
5 & Tooling / actuation       & 50 Hz    & 20 ms & 8  \\
6 & High-frequency accelerometer & 50 Hz  & 20 ms & 38 \\
\midrule
  & \textbf{Total}            &          &       & \textbf{81} \\
\bottomrule
\end{tabular}
\end{table}

\subsection{Hourly Word and Document Representation}
\label{sec:representation}

For consistency with the representation established in the underlying TQRNN methodology, the terms \emph{word} and \emph{document} are retained to denote temporal granularity. An \emph{hourly word} represents one hour of causally aligned multivariate machine observations, while a \emph{document} is the temporally ordered sequence of hourly words spanning the complete observation window. For the \TQRNNd{} configuration, each document contains $K=720$ hourly words.

The terminology defines the hierarchical organisation of the industrial time series and does not imply a linguistic input representation. Each hourly word contains continuous multivariate sensor observations rather than a discrete vocabulary element. The word-level observations are transformed by the dual-stage QRNN feature extractor into learned quantile-state representations, and the resulting document-level sequence is processed by the causal recurrent and attention mechanisms. Temporal ordering is preserved throughout the architecture, and no future observation is available during either feature extraction or classification.

\subsubsection{Multi-rate causal alignment}
\label{sec:alignment}
The six sensor groups are acquired asynchronously at three native rates: 1\,Hz,
0.033\,Hz, and 50\,Hz. For analytical consistency, the streams are projected onto
a fixed 20\,ms reference grid. Channels already sampled at 20\,ms align directly;
lower-frequency channels are lifted onto the same timeline by causal zero-order
hold, so each observed value is retained only until the next native sample becomes
available. This operation preserves temporal order and does not introduce future
measurements.

The common grid is an upstream analytical representation rather than the temporal
resolution of the long-horizon classifier. For the 81-channel configuration, one
hour contains 180,000 aligned reference steps. These samples are processed by the
upstream feature-extraction pathway and represented to the long-horizon classifier
as one 324-dimensional quantile-state vector per hourly word. A 30-day document
therefore reaches the temporal classifier as 720 ordered word-level feature vectors,
not as a 50\,Hz sequence. Native-rate modality-specific encoders or asynchronous
late fusion were not evaluated and remain alternative designs for reducing
alignment overhead.

One hour of causally aligned sensor behaviour forms one hourly word. For the
30-day observation window, the resulting document contains
\begin{equation}
K = 30 \times 24 = 720
\end{equation}
hourly words. Successive documents are refreshed at one-hour intervals and
therefore retain the temporal organisation established in the underlying
TQRNN dataset construction. The latent-history stream additionally stores a
168-hour look-back of front-end MLP embeddings, allowing the classifier to
access recent operating memory without reprocessing the complete raw trace.

Let $\mathbf{w}_{k,s}$ denote the physical-sensor component of hourly word
$k$. The trained dual-stage quantile feature extractor maps this word to
\begin{equation}
\mathbf{f}_k =
\Phi_Q(\mathbf{w}_{k,s})
\equiv \mathbf{P}(k)
\in\mathbb{R}^{324},
\label{eq:word_feature}
\end{equation}
where $\mathbf{P}(k)$ contains the four retained quantile features for each
of the 81 physical sensor channels.

The corresponding document-level quantile feature sequence is
\begin{equation}
\mathbf{F} =
[\mathbf{f}_1,\mathbf{f}_2,\ldots,\mathbf{f}_K]
\in\mathbb{R}^{K\times324},
\qquad K=720.
\label{eq:document}
\end{equation}
The temporal ordering of the hourly words is preserved throughout the
classifier. The document is therefore an ordered physical time-series
representation rather than a permutation-invariant collection.

\subsection{Label Construction}
\label{sec:labels}
Ground-truth labels are established from three independent operational sources:
operator logs, PLC fault codes, and maintenance records documenting intervention
and root cause. Domain-expert review is used for ambiguous cases. Because the
reliability of every reported classification metric depends on this procedure, it
is stated explicitly.

\paragraph{Fault-event definition}
Ground-truth fault events were established by triangulating three independent operational sources: operator logs, PLC fault codes, and maintenance records documenting intervention and root cause. An event was accepted as a confirmed failure when supported by at least two of the three sources; ambiguous cases were escalated for domain-expert review. Samples following a confirmed failure were excluded until normal operation had been restored.

\paragraph{Reference time and target window}
For an observation window ending at reference time $t_{0}$ and prediction horizon $H$, the model input is restricted to information available before $t_{0}$ and the future target interval is $(t_{0},t_{0}+H]$. The binary label is positive if at least one confirmed fault event occurs within this interval and negative otherwise. Thus,
\begin{equation}
y(t_{0},H)=
\mathbb{I}\!\left(
\exists\,t_f\in(t_{0},t_{0}+H]
\right),
\label{eq:binary_label}
\end{equation}
where $t_f$ denotes the time of a confirmed fault event.

\paragraph{Leakage control}
Strict temporal causality was maintained throughout dataset construction, such that no observations from the prediction horizon were available to the model. For the chronological short-horizon pilot, a boundary exclusion equal to the 1-hour observation window was applied at the train--validation and validation--test boundaries. The \TQRNNd{} deployment instead used machine-disjoint partitioning, with all samples from a machine assigned exclusively to training, validation, or testing.

\paragraph{Planned maintenance and data exclusions}
Planned maintenance was treated as known operational context rather than as a failure label where schedule information was available at prediction time. Sensor gaps associated with maintenance, recommissioning, or network interruptions were handled using the fixed duration-based missing-data protocol; windows intersecting gaps greater than 300\,s were excluded from model development and evaluation.

\paragraph{Label verification}
No single operational source was treated as independently definitive. Operator records, PLC fault codes, and maintenance documentation were cross-checked, with a two-of-three confirmation rule used to establish ground truth and domain-expert review applied to unresolved cases. A separate inter-annotator agreement statistic was not calculated and is therefore not reported.

\paragraph{Class prevalence}
Class prevalence was calculated at the supervised-window level rather than from individual sensor observations. The natural \TQRNNd{} partitions retained positive-class prevalences of 26.8\%, 25.9\%, and 27.3\% for training, validation, and testing, respectively, before any training-only class balancing. The increased positive-class prevalence at long horizons reflects the greater probability that a confirmed future failure falls within an extended prediction interval. Consequently, performance is evaluated using F1, recall, precision, ROC--AUC, and precision--recall measures in addition to accuracy.

\subsection{Machine-Disjoint Partitioning and Leakage Control}
\label{sec:splits}
The final \TQRNNd{} regime uses the full deployment of 72 machines across nine
manufacturing facilities. Partitioning is performed at the machine level: all
retained windows from an individual machine are assigned exclusively to one
subset. This prevents raw time-series measurements and overlapping documents from
the same machine from being shared between training, validation, and testing.

The allocation uses 43 machines (59.7\%)
for training, 14 machines (19.4\%) for validation, and 15 machines (20.8\%) for
testing. Allocation is stratified across the nine participating facilities so that
each partition contains machines from the observed site population. This is a
machine-disjoint, multi-site held-out test; it is not a leave-one-site-out test and
does not establish transfer to a facility absent from model development.

All document construction remains causal. For an observation document ending at
$t_0$, only information available before $t_0$ enters the input and the target
interval lies strictly in the future. Preprocessing quantities, including
normalisation statistics, imputation rules, class weights, and model-selection
parameters, are estimated from the applicable training/development data and then
held fixed for validation and held-out evaluation. The earlier short-horizon pilot
used its own chronological split; those regime-specific procedures are not
retroactively attributed to the 72-machine long-horizon results reported here.

\section{Proposed Framework}
\label{sec:method}

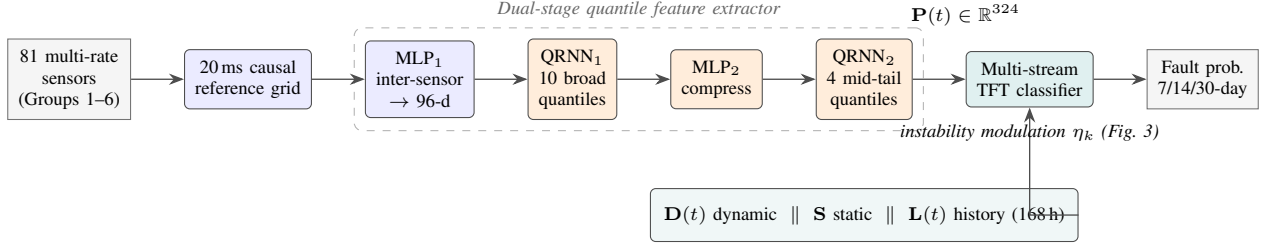
\begin{figure*}[t]
\centering
\resizebox{\ifdim\width>\textwidth\textwidth\else\width\fi}{!}{%
\begin{tikzpicture}[node distance=6mm and 7mm]
\node[io] (sens) {81 multi-rate\\sensors\\(Groups 1--6)};
\node[stage, right=of sens] (grid) {20\,ms causal\\reference grid};
\node[stage, right=of grid] (mlp1) {MLP$_1$\\inter-sensor\\$\rightarrow$ 96-d};
\node[qr, right=of mlp1] (qr1) {QRNN$_1$\\10 broad\\quantiles};
\node[qr, right=of qr1] (mlp2) {MLP$_2$\\compress};
\node[qr, right=of mlp2] (qr2) {QRNN$_2$\\4 mid-tail\\quantiles};
\node[cls, right=of qr2] (tft) {Multi-stream\\TFT classifier};
\node[io, right=of tft] (out) {Fault prob.\\7/14/30-day};
\foreach \a/\b in {sens/grid,grid/mlp1,mlp1/qr1,qr1/mlp2,mlp2/qr2,qr2/tft,tft/out}
  \draw[ar] (\a) -- (\b);
\node[font=\scriptsize\itshape, above=0.5mm of qr2.north east, anchor=south, xshift=7mm]
   {$\mathbf{P}(t)\in\mathbb{R}^{324}$};
\node[block=teal!6, below=9mm of qr2, inner sep=5pt] (side)
   {$\mathbf{D}(t)$ dynamic $\;\Vert\;$ $\mathbf{S}$ static $\;\Vert\;$ $\mathbf{L}(t)$ history (168\,h)};
\draw[ar] (side.east) -| (tft.south);
\node[font=\scriptsize\itshape, below=1mm of tft.south, anchor=north]
   {instability modulation $\eta_k$ (Fig.~\ref{fig:classifier})};
\begin{scope}[on background layer]
\node[grp, fit=(mlp1)(qr1)(mlp2)(qr2),
      label={[font=\scriptsize\itshape,black!60]above:Dual-stage quantile feature extractor}] {};
\end{scope}
\end{tikzpicture}}
\caption{Overview of the \TQRNNd{} framework. Multi-rate 81-channel sensor data
are aligned onto a 20\,ms causal reference grid and segmented into 720 hourly
words. A dual-stage QRNN produces a 324-dimensional mid-tail quantile-state
representation $\mathbf{P}(t)$, which is fused with dynamic covariates
$\mathbf{D}(t)$, static metadata $\mathbf{S}$, and a 168-hour latent-history
summary $\mathbf{L}(t)$ inside a multi-stream temporal-fusion classifier that
outputs structured forecast-interval probabilities for the 7-, 14-, and 30-day horizons.}
\label{fig:pipeline}
\end{figure*}

\TQRNNd{} consists of two main components: a dual-stage QRNN feature extractor
and a multi-stream temporal-fusion classifier. The extractor learns a compact
distribution-aware representation of each hourly word. The classifier
then fuses this representation with metadata, dynamic covariates, and historical
memory to estimate the probability of abnormal behaviour at future horizons.
Fig.~\ref{fig:pipeline} shows the complete pipeline and the point at which each
information stream enters; the two components are described in
Sections~\ref{sec:twostage} and~\ref{sec:classifier} respectively, and the
auxiliary gate in Section~\ref{sec:gate}.

\subsection{Why a Quantile Encoding Rather Than Point Summaries}
\label{sec:whyquantiles}
The feature-extraction design is motivated by the observation that pre-failure
behaviour is not expressed only as a shift in mean sensor level. In the underlying
industrial study, degradation may appear as widening dispersion, asymmetric tails,
or changes in the relative position of observations within the operating
distribution. Point-estimate summaries can obscure these effects through averaging,
whereas conditional quantiles preserve distributional structure explicitly.

Three properties are particularly relevant. First, quantile representations expose
changes in distributional spread before a dominant central-location shift is
visible. Second, separate upper and lower quantiles retain directional information
about asymmetric movement; for example, the pair $q_{0.25}$ and $q_{0.75}$
preserves the interquartile envelope rather than collapsing it into one symmetric
moment. Third, quantile-based features remain useful in the heavy-tailed and
occasionally impulsive operating distributions encountered in production, where a
single Gaussian description is often inappropriate. These properties make the
quantile representation a suitable intermediate interface between heterogeneous
sensor data and downstream temporal classification.

The long-horizon second-stage set
$\{0.25,0.40,0.60,0.75\}$ is a methodological refinement of the original
$\{0.25,0.75\}$ representation rather than the result of an exhaustive search over
candidate level sets. The 0.25 and 0.75 levels preserve the interquartile envelope,
while 0.40 and 0.60 add an interior pair around the median so that upward and
downward displacement around the distribution centre remains distinguishable. The
median itself remains represented in the broader first-stage ten-quantile set, as
do the more extreme quantiles. The purpose of QRNN$_2$ is therefore targeted
second-stage estimation in the intermediate region rather than duplication of a
subset of QRNN$_1$ outputs.

The day-scale ablation in Table~\ref{tab:quantile_sens} provides the
empirical evidence for the representation hierarchy. At 30 days, the
Transformer-only day-scale configuration reaches 67.13\% F1, whereas the fully
integrated two-stage quantile pipeline reaches 79.97\%. The isolated QRNN stages
are weaker than the integrated model, indicating that the strongest result arises
from coupling distribution-aware feature extraction with long-context temporal
classification rather than from either block in isolation.

\subsection{Dual-Stage Quantile Feature Extraction}
\label{sec:twostage}
\begin{figure}[t]
\centering
\resizebox{\ifdim\width>\columnwidth\columnwidth\else\width\fi}{!}{%
\begin{tikzpicture}[node distance=4.5mm and 6mm]
\node[io] (in) {Aligned 81-ch word\\+ temporal embedding};
\node[stage, below=of in] (m1) {MLP$_1\;\rightarrow\;$96-d};
\node[qr, below=of m1] (q1) {QRNN$_1$: broad set\\$\alpha\in\{0.01,\dots,0.99\}$\\($10\times81=810$)};
\node[qr, below=of q1] (m2) {MLP$_2$ compression};
\node[qr, below=of m2] (q2) {QRNN$_2$: mid-tail\\$\alpha\in\{0.25,0.40,0.60,0.75\}$};
\node[cls, below=of q2] (p) {$\mathbf{P}(t)\in\mathbb{R}^{324}$};
\foreach \a/\b in {in/m1,m1/q1,q1/m2,m2/q2,q2/p} \draw[ar] (\a) -- (\b);
\end{tikzpicture}}
\caption{Dual-stage quantile feature extractor. QRNN$_1$ learns the broad
conditional distribution of each channel; MLP$_2$ compresses it; and QRNN$_2$
refines the four mid-tail quantiles used as the classifier-facing
representation $\mathbf{P}(t)$.}
\label{fig:qrnn}
\end{figure}
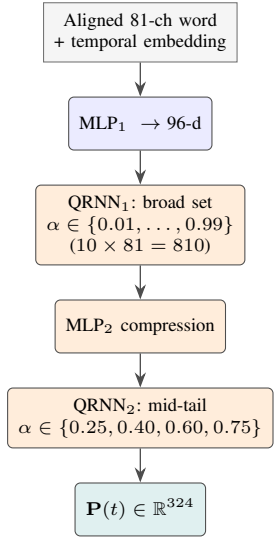

The feature extractor, summarised in Fig.~\ref{fig:qrnn}, transforms each hourly
word into a compact mid-tail quantile representation through two cascaded
quantile-regression stages.

\paragraph{Why two stages rather than one}
The two-stage hierarchy separates broad distribution learning from targeted
intermediate-quantile refinement. QRNN$_1$ estimates ten conditional quantiles
spanning $\alpha\in[0.01,0.99]$, giving the upstream representation access to the
wider conditional distribution. MLP$_2$ then compresses this representation before
QRNN$_2$ estimates the four retained intermediate levels. The purpose of the
second stage is therefore not to repeat the first-stage outputs, but to concentrate
representational capacity on the region used by the long-horizon classifier while
preserving information inherited from the broader distributional encoding.

\paragraph{Retained second-stage quantiles}
The four-level set $\{0.25,0.40,0.60,0.75\}$ extends the original two-quantile
interquartile representation used in the earlier TQRNN pipeline. The 0.25 and 0.75
levels preserve that envelope, while 0.40 and 0.60 form an additional symmetric
interior pair around the conditional median. QRNN$_1$ continues to estimate
$\alpha=0.50$ and the outer quantiles as part of its ten-level set. No exhaustive
candidate-set search or fold-stability claim is made for the four retained levels.

\begin{table*}[t]
\centering
\caption{Long-horizon feature-extraction/pipeline ablation at 7, 14, and 30 days. P9 is the transferred 70-hour reference; P10 and P11 isolate the first and second quantile stages; P12 is the Transformer-only day-scale model; P13 is the fully integrated \TQRNNd{} pipeline. Values are percentages.}
\label{tab:quantile_sens}
\scriptsize
\setlength{\tabcolsep}{3.2pt}
\resizebox{\textwidth}{!}{%
\begin{tabular}{@{}lcccccccccccc@{}}
\toprule
& \multicolumn{4}{c}{\textbf{7-day}} & \multicolumn{4}{c}{\textbf{14-day}} & \multicolumn{4}{c}{\textbf{30-day}}\\
\cmidrule(lr){2-5}\cmidrule(lr){6-9}\cmidrule(lr){10-13}
\textbf{Configuration} & F1 & Rec. & Prec. & Acc. & F1 & Rec. & Prec. & Acc. & F1 & Rec. & Prec. & Acc.\\
\midrule
All$_{70\mathrm{h}}$ & 61.49 & 61.08 & 62.27 & 62.55 & 57.58 & 57.11 & 56.59 & 57.37 & 52.41 & 53.70 & 53.30 & 53.72\\
QRNN$_1^{30\mathrm{d}}$ & 56.59 & 56.95 & 57.85 & 58.48 & 52.44 & 53.21 & 53.42 & 53.54 & 50.60 & 49.87 & 51.65 & 52.06\\
QRNN$_2^{30\mathrm{d}}$ & 61.81 & 62.62 & 63.33 & 63.39 & 58.68 & 59.45 & 59.81 & 59.62 & 56.17 & 56.77 & 57.27 & 57.21\\
Transformer$_{30\mathrm{d}}$ & 72.56 & 72.81 & 72.92 & 71.93 & 66.52 & 65.19 & 66.26 & 65.76 & 67.13 & 66.54 & 66.47 & 67.51\\
\textbf{All$_{30\mathrm{d}}$} & \textbf{76.09} & \textbf{75.81} & \textbf{76.89} & \textbf{77.14} & \textbf{76.76} & \textbf{78.73} & \textbf{78.31} & \textbf{78.69} & \textbf{79.97} & \textbf{80.18} & \textbf{81.82} & \textbf{82.39}\\
\bottomrule
\end{tabular}}
\end{table*}

A front-end multilayer perceptron, MLP$_1$, first encodes cross-sensor
relationships from the aligned 81-channel word and its temporal embedding into a
96-dimensional latent projection. The 96-dimensional width is the implemented
long-horizon configuration; no separate projection-width sensitivity study is claimed. The first QRNN
stage estimates a broad set of quantiles,
\[
\alpha \in \{0.01,0.10,0.20,0.25,0.50,0.60,0.75,0.80,0.90,0.99\},
\]
capturing the wider conditional distribution of each channel. A second MLP
compresses this representation, and the second QRNN stage refines the
mid-tail quantiles
\[
\alpha \in \{0.25,0.40,0.60,0.75\},
\]
which are used as the classifier-facing representation.

For predicted quantile $\hat{q}_{\alpha}(x_{t,i})$ and observation $y_{t,i}$,
the pinball loss is
\begin{equation}
\ell_{\alpha}(y_{t,i},\hat{q}_{\alpha}) =
\max\!\left(\alpha(y_{t,i}-\hat{q}_{\alpha}),
(\alpha-1)(y_{t,i}-\hat{q}_{\alpha})\right).
\label{eq:pinball}
\end{equation}
The loss is summed over channels, time steps, and quantile levels.

The refined quantile-state vector at hourly word $t$ is
\begin{equation}
\mathbf{P}(t)=
\operatorname{vec}\!\left(
[q^{(i)}_{0.25},q^{(i)}_{0.40},q^{(i)}_{0.60},q^{(i)}_{0.75}]_{i=1}^{81}
\right)\in\mathbb{R}^{324}.
\label{eq:quantile_state}
\end{equation}
The resulting 30-day document-level quantile feature sequence is therefore
$\mathbf{F}\in\mathbb{R}^{720\times324}$.

\subsection{Multi-Stream Temporal-Fusion Classifier}
\label{sec:classifier}
The long-horizon classifier uses four information streams. The TFT components---
gated residual processing, variable selection, causal recurrent encoding, and
attention---follow established temporal-fusion principles \cite{Lim2021}; the
specific interface in this work is the coupling of the QRNN-derived document
representation with dynamic, static, and historical context.

\begin{itemize}
    \item \textbf{Quantile state} $\mathbf{P}(t)\in\mathbb{R}^{324}$, formed from
    four retained quantiles for each of the 81 physical sensor channels.
    \item \textbf{Dynamic covariates} $\mathbf{D}(t)\in\mathbb{R}^{29}$,
    comprising environmental, facility, operational, and known-ahead contextual
    variables available at prediction time; these are projected to 516 dimensions.
    \item \textbf{Static channel metadata} $\mathbf{S}_i\in\mathbb{R}^{3}$ for
    channel $i$, encoding channel type, physical location, and the causally
    available channel-health/degradation-prior indicator. Each channel vector is
    embedded as $\mathbf{S}'_i\in\mathbb{R}^{64}$ and collected in
    \begin{equation}
    \mathbf{S}_{\mathrm{bank}}=
    [\mathbf{S}'_1;\ldots;\mathbf{S}'_{C_s}]
    \in\mathbb{R}^{81\times64}.
    \label{eq:sbank}
    \end{equation}
    An aggregated 64-dimensional static representation $\mathbf{S}'$ is used for
    global conditioning, whereas $\mathbf{S}_{\mathrm{bank}}$ is retained for
    channel-indexed cross-modal attention.
    \item \textbf{Historical latent memory} $\mathbf{L}(t)$, formed from stored
    96-dimensional MLP$_1$ embeddings over a 168-hour look-back and projected to
    192 dimensions before recurrent fusion.
\end{itemize}

At hourly word position $t$, the temporally varying streams form
\begin{equation}
\mathbf{e}_{t}=
[\mathbf{P}'(t)\Vert\mathbf{D}'(t)\Vert\mathbf{L}'(t)]
\in\mathbb{R}^{1032},
\label{eq:fused_input}
\end{equation}
where $1032=324+516+192$. The static representations remain separate
conditioning inputs and do not increase this recurrent width.

Gated residual networks provide nonlinear transformation with residual and gated
information flow. In compact form,
\begin{equation}
\begin{aligned}
u &= W_1\,\mathrm{ELU}(W_2x+b_2)+b_1,\\
\mathrm{GRN}(x) &= \mathrm{LayerNorm}(x+\mathrm{GLU}(u)).
\end{aligned}
\label{eq:grn}
\end{equation}
Context-conditioned variable selection uses the aggregated static representation
$\mathbf{S}'$ to reweight the temporally varying inputs before a causal GRU encoder.
The recurrent pathway processes the document strictly in temporal order; no
future word is available when computing the hidden state at the current position.

After causal recurrent encoding, cross-modal attention conditions the evolving
temporal evidence on channel-indexed static metadata. Let $\mathbf{E}(t)$ denote
the encoded temporal-evidence representation presented to the attention block. For
attention head $m$, the temporal pathway provides the query and the metadata bank
provides keys and values:
\begin{equation}
\begin{aligned}
\mathbf{q}^{(m)}(t) &= \mathbf{E}(t)W_Q^{(m)},\\
\mathbf{K}^{(m)}_{\mathrm{att}} &= \mathbf{S}_{\mathrm{bank}}W_K^{(m)},\\
\mathbf{A}^{(m)}(t) &=
\mathrm{softmax}\!\left(
\frac{\mathbf{q}^{(m)}(t)(\mathbf{K}^{(m)}_{\mathrm{att}})^\top}
{\sqrt{d_k}}
\right),
\end{aligned}
\label{eq:attention}
\end{equation}
where the softmax is taken over the $C_s=81$ channel-metadata entries. With a
shared value projection
$\mathbf{V}_{\mathrm{att}}=\mathbf{S}_{\mathrm{bank}}W_V$, the attended metadata
context is obtained from the head-aggregated weights and is fused with the temporal
pathway before classification.

For prediction horizon $H$, the document-level classification head produces
$K_H$ hourly forecast-interval logits, where $K_H\in\{168,336,720\}$ for the
7-, 14-, and 30-day tasks. Element-wise sigmoid activation gives
\begin{equation}
\hat{\mathbf{y}}=
[\hat{y}_1,\ldots,\hat{y}_{K_H}]\in[0,1]^{K_H},
\label{eq:structured_output}
\end{equation}
with $\hat{y}_r$ denoting the estimated failure probability in forecast interval
$I_r$. The headline Normal/Abnormal decision is obtained by thresholding the
structured output at the selected operating point; the interval structure is
retained internally rather than replaced by a scalar RUL target.

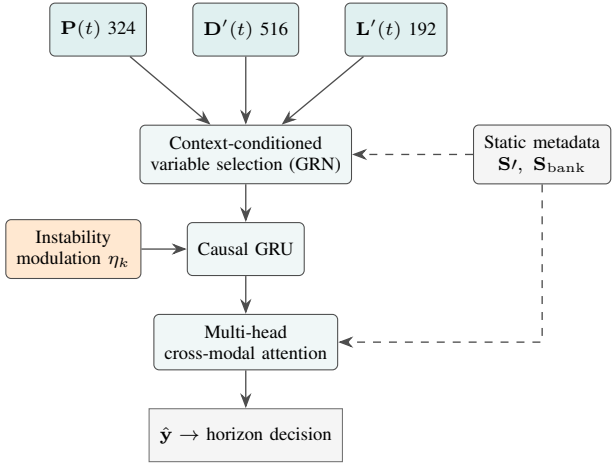
\begin{figure}[t]
\centering
\resizebox{\ifdim\width>\columnwidth\columnwidth\else\width\fi}{!}{%
\begin{tikzpicture}[node distance=5mm and 6mm]
\node[cls] (p) {$\mathbf{P}(t)$ 324};
\node[cls, right=of p] (d) {$\mathbf{D}'(t)$ 516};
\node[cls, right=of d] (l) {$\mathbf{L}'(t)$ 192};
\node[block=teal!6, below=9mm of d] (vs) {Context-conditioned\\variable selection (GRN)};
\node[block=teal!6, below=of vs] (gru) {Causal GRU};
\node[block=orange!18, left=of gru] (gate) {Instability\\modulation $\eta_k$};
\node[block=teal!6, below=of gru] (att) {Multi-head\\cross-modal attention};
\node[io, below=of att] (out) {$\hat{\mathbf{y}}\rightarrow$ horizon decision};
\node[block=black!4, right=16mm of vs] (s) {Static metadata\\$\mathbf{S}\prime,\;\mathbf{S}_{\rm bank}$};
\foreach \a in {p,d,l} \draw[ar] (\a) -- (vs);
\draw[ar] (vs) -- (gru);
\draw[ar] (gate) -- (gru);
\draw[ar] (gru) -- (att);
\draw[ar] (att) -- (out);
\draw[ar, dashed] (s) -- (vs);
\draw[ar, dashed] (s.south) |- (att.east);
\end{tikzpicture}}
\caption{Multi-stream temporal-fusion classifier. The quantile state, dynamic
covariates, and historical-latent streams are reweighted by
metadata-conditioned variable selection, encoded by a causal GRU whose update
gate is modulated by the bounded instability signal $\eta_k$, and fused
by metadata-conditioned cross-modal attention. Dashed arrows denote static-
metadata conditioning; the final head emits structured hourly forecast-interval
probabilities before the horizon-level decision is formed.}
\label{fig:classifier}
\end{figure}

\subsection{Instability-Aware Memory Gate}
\label{sec:gate}
For the longest-horizon pathway, \TQRNNd{} uses an auxiliary instability-aware
signal to support temporal memory retention when local prediction-error divergence
persists across successive hourly words. The mechanism is deliberately limited: it
is not an estimate of a global or local Lyapunov exponent, does not perform online
phase-space reconstruction, and is not a separate fault classifier.

For sensor channel $i$, let $u_i(k)$ denote the observed auxiliary channel-state
summary at hourly word $k$, derived from the retained channel-level quantile-state
representation. Let $\hat{u}_i(k\mid<k)$ denote its one-word-ahead prediction using
only information available strictly before word $k$. The auxiliary prediction error
is
\begin{equation}
e_i(k)=u_i(k)-\hat{u}_i(k\mid<k).
\label{eq:aux_error}
\end{equation}
The local prediction-error divergence score is
\begin{equation}
\lambda_i(k)=
\log\!\left(
\frac{\lVert e_i(k)-e_i(k-1)\rVert+\epsilon}
{\lVert e_i(k-1)-e_i(k-2)\rVert+\epsilon}
\right),
\label{eq:lambda}
\end{equation}
with $\epsilon=10^{-3}$. A channel is treated as exhibiting sustained local
divergence only after three consecutive threshold exceedances:
\begin{equation}
I_i(k)=\mathbb{I}\!\left[
\lambda_i(k)>\lambda_0\wedge
\lambda_i(k-1)>\lambda_0\wedge
\lambda_i(k-2)>\lambda_0
\right],
\label{eq:indicator}
\end{equation}
where $\lambda_0=0.2$ in the evaluated configuration. For the degradation-
sensitive channel subset $\mathcal{C}$, the aggregate signal is
\begin{equation}
\Lambda_{\mathcal{C}}(k)=
\sum_{i\in\mathcal{C}}w_i I_i(k)\lambda_i^{+}(k),
\qquad
\lambda_i^{+}(k)=\max(0,\lambda_i(k)),
\label{eq:aggregate}
\end{equation}
where $w_i$ are learned non-negative channel weights.

Before entering the temporal gate, the aggregate score is bounded:
\begin{equation}
\eta_k=\sigma\!\left(a_{\Lambda}\Lambda_{\mathcal{C}}(k)+b_{\Lambda}\right),
\qquad \eta_k\in(0,1),
\label{eq:bounded_gate}
\end{equation}
with learned scalar parameters $a_{\Lambda}$ and $b_{\Lambda}$. The instability-
aware update gate is then
\begin{equation}
\mathbf{z}^{\Lambda}_k=
\sigma\!\left(
W_{xz}\mathbf{x}^{h}_k+W_{hz}\mathbf{h}_{k-1}+\mathbf{b}_z
+\kappa_z\eta_k\mathbf{1}_{d_h}
\right),
\label{eq:modulated_gate}
\end{equation}
where $\kappa_z\ge0$ controls the strength of the bounded memory bias. Under
sustained local divergence, the additive term biases the update gate towards
retaining more of the previous hidden state. The retained QRNN representation
remains the primary predictive basis; the instability pathway supplies only an
auxiliary modulation signal. Its independent performance contribution was not
isolated in a dedicated with/without ablation in the completed experimental
programme and is therefore not claimed separately.

\subsubsection{Interpretation and limitations of instability-aware gating}
\label{sec:gate_confound}

The instability-aware gate is designed to identify sustained local divergence in the prediction-error trajectory rather than to classify its physical cause. Consequently, a persistent increase in the auxiliary divergence score may arise from progressive degradation, but may also be produced by sensor drift, material or operating-regime transitions, or other sustained departures from the learned operating distribution. The gate should therefore be interpreted as a temporal modulation signal and not as an independent fault detector.

Two design features reduce sensitivity to non-degradation disturbances. First, the aggregate instability measure in (\ref{eq:aggregate}) is restricted to the degradation-sensitive channel subset $\mathcal{C}$, limiting the contribution of channels dominated by contextual or instrumentation-related variation. Second, the resulting instability signal is passed through a bounded learned modulation pathway, allowing its influence on temporal memory to be attenuated when it is not informative. The persistence criterion in (\ref{eq:indicator}) further suppresses isolated transient excursions by requiring consecutive threshold exceedances, although sustained non-fault regime changes may still activate the pathway.

This limitation is particularly relevant during material, lubrication, or operating-condition transitions, where sensor distributions may shift without an associated mechanical failure. Accordingly, the instability-aware pathway is used only as an auxiliary contribution to the \TQRNNd{} representation; the final classification decision remains determined by the complete quantile-derived, temporal, contextual, and historical feature set.

\subsubsection{Instability-gate operating configuration}
\label{sec:gatesens}

The auxiliary instability pathway is governed by the empirical divergence threshold $\lambda_{0}$, the temporal persistence requirement, the degradation-sensitive channel subset $\mathcal{C}$, and learned modulation parameters. The implemented configuration uses $\lambda_{0}=0.2$ with activation requiring three consecutive hourly threshold exceedances. A stabilising constant of $\epsilon=10^{-3}$ is applied in the local divergence calculation, and the retained quantile context is $\alpha={0.25,0.40,0.60,0.75}$.

These quantities are implementation parameters rather than physical constants. Threshold selection was confined to the training and validation partitions, with the held-out test data excluded from parameter selection. The purpose of the gate is therefore not to establish a physically unique measure of degradation, but to provide a bounded auxiliary indication of sustained instability that can strengthen long-horizon temporal retention when supported by the wider learned representation.

\begin{table}[t]
\centering
\caption{Implemented instability-aware gating configuration for \TQRNNd{}.}
\label{tab:gate_sens}
\footnotesize
\setlength{\tabcolsep}{3pt}
\begin{tabular}{@{}p{0.29\columnwidth}p{0.25\columnwidth}p{0.36\columnwidth}@{}}
\toprule
\textbf{Quantity} & \textbf{Value} & \textbf{Role}\\
\midrule
Cadence & 1 hourly word & Aligns divergence scoring with the hourly word cadence of each document.\\
Stabilising constant $\epsilon$ & $10^{-3}$ & Prevents zero/near-zero denominator in the local error-change ratio.\\
Divergence threshold $\lambda_0$ & 0.2 & Empirical operating threshold; approximately a 1.22 error-change ratio before persistence is applied.\\
Confirmation rule & 3 consecutive hours & Rejects isolated noise spikes and short transients.\\
Retained quantiles & $\{0.25,0.40,\allowbreak 0.60,0.75\}$ & Keeps the gate aligned with the retained long-horizon quantile surface.\\
Channel set $\mathcal{C}$ & Degradation-sensitive subset & Restricts aggregation to channels relevant to long-horizon degradation.\\
\bottomrule
\end{tabular}
\end{table}

The implementation uses the operating configuration in
Table~\ref{tab:gate_sens}. The threshold and persistence settings were selected
using development data and then held fixed for held-out evaluation. No claim is
made that these values are physically unique or transferable without recalibration.

\subsection{Training Objective}
The two QRNN stages are trained using the pinball quantile loss in
(\ref{eq:pinball}) to construct the upstream quantile representation. The
long-horizon TFT classifier is then trained on that QRNN-derived representation as
a document-level, multi-output binary classifier. For training document $b$ and
forecast interval $r$, let $y_{b,r}\in\{0,1\}$ and
$\hat{y}_{b,r}\in[0,1]$ denote the target and predicted failure probability. The
interval-wise binary cross-entropy is
\begin{equation}
\mathcal{L}_{r}=
-\left[y_{b,r}\log\tilde{y}_{b,r}
+(1-y_{b,r})\log(1-\tilde{y}_{b,r})\right],
\label{eq:bce}
\end{equation}
where $\tilde{y}_{b,r}=\mathrm{clip}(\hat{y}_{b,r},\varepsilon,1-\varepsilon)$
for numerical stability.

Because confirmed-failure intervals are sparse relative to non-failure intervals,
the implemented classifier uses class-weighted BCE where appropriate:
\begin{equation}
\mathcal{L}^{\mathrm{wBCE}}_{r}=
-\left[\omega_{+}y_{b,r}\log\tilde{y}_{b,r}
+\omega_{-}(1-y_{b,r})\log(1-\tilde{y}_{b,r})\right].
\label{eq:wbce}
\end{equation}
The weights are calculated from the training partition only using inverse class
prevalence and are held fixed during validation and testing. For mini-batch
$\mathcal{B}$, the classifier objective is
\begin{equation}
\mathcal{L}_{\mathrm{TFT}}(\theta)=
\frac{1}{|\mathcal{B}|K_H}
\sum_{b\in\mathcal{B}}\sum_{r=1}^{K_H}
\mathcal{L}^{\mathrm{wBCE}}_{r}
+\rho_{\mathrm{wd}}\lVert\theta\rVert_2^2,
\label{eq:total_loss}
\end{equation}
where $\rho_{\mathrm{wd}}$ is the weight-decay coefficient reported in
Table~\ref{tab:hyperparams}. The decision threshold is applied only at inference
time; it is not part of the training loss. Validation-based early stopping and the
learning-rate schedule in Table~\ref{tab:hyperparams} are used during classifier
training.

\section{Experimental Setup}
\label{sec:setup}

\subsection{Baselines and Comparison Protocol}
\label{sec:baselines}

\subsubsection{Baseline selection and coverage}
The completed comparison contains 18 baselines spanning complementary model
families rather than multiple near-duplicates of one architecture. The purpose is
to test whether the observed long-horizon performance ordering persists against
kernel, instance-based, tree, recurrent, feed-forward, reconstruction, attention,
and pretrained-transformer alternatives under the same regime-specific data and
evaluation foundation.

\begin{itemize}
\item \textbf{Kernel and instance-based} (support-vector regression, SVR;
least-squares support-vector machine, LS-SVM; $k$-nearest neighbours, KNN):
test whether the
task is separable without temporal modelling at all.
\item \textbf{Tree ensembles} (XGBoost, LightGBM): widely used tabular alternatives for non-sequential industrial features.
\item \textbf{Recurrent} (long short-term memory, LSTM; long short-term recurrent network, LSRN; liquid time-constant network, Liquid): explicit sequential memory,
the standard comparator in the PdM literature \cite{Meddaoui,Mitici}.
\item \textbf{Feed-forward} (MLP): an ablation of temporal structure at fixed
capacity.
\item \textbf{Generative and reconstruction-based} (GAN, Autoencoder): the
anomaly-detection inductive bias, included to test the alternative framing
discussed in Section~\ref{sec:rw_anomaly}.
\item \textbf{Attention} (Transformer, Transformer-XL, TFT$_{\text{full}}$):
long-range attention, with TFT$_{\text{full}}$ the direct architectural
ancestor and hence the most informative single comparator.
\item \textbf{Pretrained language-model architectures} (BERT, RoBERTa,
DistilBERT, ALBERT): included because the sequence-classification framing of
Section~\ref{sec:representation} invites the question of whether a pretrained
sequence classifier suffices.
\end{itemize}

\subsubsection{Adaptation of BERT-family sequence baselines}
The BERT, RoBERTa, DistilBERT, and ALBERT rows are retained from the completed
comparison as sequence-model baselines. They test whether the observed
performance ordering can be explained simply by adopting a large pretrained
transformer-family architecture. The reported comparison establishes their scores
under the same long-horizon task and held-out population, but the retained experimental
record does not provide sufficient adapter detail to justify a new token-level or
positional-embedding specification in this article. No such implementation detail
is therefore reconstructed retrospectively; independent code-level replication
would require the archived experiment artefacts.

\subsubsection{Controlled comparison protocol}
\label{sec:matched_comparison}

Comparative fairness was enforced at the data, task, and evaluation levels.
For each forecast horizon, all models use the same machine-disjoint
training, validation, and held-out test assignments. Preprocessing quantities and
input transformations are estimated from the applicable training/development data
only and then held fixed for evaluation.

All models were trained for the same binary fault-occurrence target and
evaluated on the same held-out instances using the same metrics. Where required, architecture-specific input adapters were used so that temporal
models received the ordered quantile-state sequence and non-sequential models
received a fixed representation derived from the same upstream feature basis. The
BERT-family rows are interpreted as architecture-family comparators as described
above; no unsupported tokenisation detail is inferred retrospectively.

Hyperparameters, stopping criteria, class-imbalance treatment, and decision
thresholds were selected using training or validation data only. Once selected,
the decision threshold was fixed before evaluation on the held-out test
partition. Because the compared model families have substantially different
optimisation requirements, no claim is made that they used identical search
spaces or equal numbers of hyperparameter trials. The matched conditions
therefore refer to the common dataset partitions, target definition, forecast
horizons, leakage controls, and held-out evaluation protocol.

\subsubsection{Scope of the reported experiments}
The evidence reported in this article comprises the completed 7-, 14-, and 30-day
cross-model benchmark, the day-scale pipeline ablation, the TFT-side flat-input
ablation, ROC--AUC analysis, precision--recall profile analysis, and computational-
cost comparison. All use the regime-specific held-out evaluation foundation described
in Section~\ref{sec:splits}. Rolling-origin evaluation, leave-one-site-out transfer,
additional long-sequence baselines, and matched one-component ablations are not
reported because they were not part of the completed experimental programme.

\subsubsection{Naming of TFT configurations}
Two distinct input regimes of the same TFT architecture appear in this paper.
They are named explicitly and are used consistently throughout:
\begin{itemize}
\item \textbf{TFT$_{\text{full}}$} --- the TFT architecture receiving the
complete MLP--QRNN quantile-state representation, as in
Table~\ref{tab:comparison_multi}. This is the cross-model benchmark.
\item \textbf{TFT$_{\text{flat}}$} --- the same architecture receiving raw
flat input with the MLP--QRNN hierarchy removed, used only as the starting
point of the ablation in Table~\ref{tab:ablation}.
\end{itemize}
These are the same network under different input regimes and their values are
not expected to match.

\subsection{Metrics}
\label{sec:metrics}
Performance is reported using F1-score, recall, precision, and accuracy at the
selected operating threshold. Unless otherwise stated, these metrics are first
computed for each evaluation unit and are then reported as unweighted means across
the corresponding held-out cohort. Consequently, the reported F1-score is the mean
of the per-unit F1-scores and is not obtained by taking the harmonic mean of the
reported mean precision and mean recall. This definition explains why the aggregate
F1 value need not equal $2PR/(P+R)$ when $P$ and $R$ denote the tabled cohort means.

Threshold-independent discrimination is assessed using ROC--AUC. Positive-class
behaviour under imbalance is additionally summarised using the mean multi-site
precision--recall profile area from the completed evaluation. This is a
profile-level area summary and is not relabelled as raw-score average precision.
Computational cost is reported using training time per epoch, inference latency,
parameter count, and peak GPU memory.

\subsection{Implementation}

Data processing and evaluation were implemented in Python using NumPy, pandas,
and SciPy. Neural-network models were implemented in PyTorch, while classical
machine-learning baselines and standard evaluation procedures used
scikit-learn. The neural models were trained using Adam with learning-rate
scheduling, validation-based early stopping, and architecture-specific dropout
and normalisation.

The \TQRNNd{} configuration uses 81 physical sensor channels, a 720-hour
input sequence, and a 168-hour latent-history window. Its temporal evidence
stream has dimension 1032, comprising the 324-dimensional QRNN representation,
516-dimensional dynamic-context projection, and 192-dimensional latent-history
projection; static metadata are embedded separately.

Training was performed using a single NVIDIA A100 GPU with 80\,GB memory,
supported by AMD EPYC 7742 and 7763 processors. Training the final dual-stage
QRNN feature extractor required approximately 20 hours, followed by
approximately 32 hours for the TFT classifier, giving approximately 52 hours
for one complete \TQRNNd{} training cycle. This figure excludes baseline and
ablation training.

Each reported model configuration corresponds to one completed training and
held-out evaluation run under its fixed experimental settings. The headline
results are therefore not averages over repeated random initializations. A systematic multi-seed retraining study was not performed, and seed-to-seed
variance is consequently not reported.

\subsection{Modelling Assumptions}
\label{sec:assumptions}
The following assumptions and limitations condition interpretation of the reported
results.
\begin{enumerate}
\item \textbf{Label reliability.} Operator records, PLC fault codes, and maintenance
records are treated as the operational ground-truth sources under the two-of-three
confirmation rule. Unrecorded faults or unlogged interventions can therefore create
label noise.
\item \textbf{Sensor validity.} Channels are assumed to remain correctly calibrated
and mapped to the intended assets over the analysed periods; the classifier is not
preceded by a separate sensor-fault diagnosis model.
\item \textbf{Distribution stability within the observed deployment.} Training-derived
normalisation and preprocessing quantities are assumed to remain applicable to the
held-out machines and periods drawn from the same industrial estate. Longer-term
concept drift is not eliminated by this design.
\item \textbf{Homogeneity of the asset family.} Pooling across facilities assumes that
the common machine family shares sufficiently similar degradation physics
and sensor semantics. This does not imply transfer to an unseen facility or another
machine class.
\item \textbf{Dependence of overlapping documents.} The machine-disjoint split prevents
raw measurements from the same machine appearing in more than one partition, but
successive 30-day documents within a machine overlap by 719 hours. Such documents are
valid prediction instances but are not interpreted as independent machines, sites, or
experimental replications.
\end{enumerate}
No synthetic data or simulation is used to generate the headline classifier
results; they derive from recorded operational data.

\subsection{Hyperparameter Configuration}
Hyperparameters for the TFT were selected for model optimisation and long-horizon
stability, ensuring robust handling of temporal and contextual data.
Table~\ref{tab:hyperparams} outlines the architectural and training configuration.

\begin{table*}[t]
\centering
\caption{TFT architecture and training hyperparameters.}
\label{tab:hyperparams}
\footnotesize
\setlength{\tabcolsep}{4pt}
\begin{minipage}[t]{0.49\textwidth}
\centering
\begin{tabular}{@{}l l l@{}}
\toprule
\textbf{Component} & \textbf{Parameter} & \textbf{Value} \\
\midrule
Primary QRNN     & Input dimension       & 324 ($4{\times}81$ quantiles) \\
stream            & Recurrent-stream dim. & 324 \\
\midrule
External context  & Raw dimension         & 29 \\
                  & Projection dimension  & 516 \\
                  & Stream role           & Direct TFT (no QRNN) \\
\midrule
Static metadata   & Raw dimension         & 3 \\
                  & Embedding dimension   & 64 \\
\midrule
Historical latent & Raw dimension         & $168{\times}96$ \\
                  & Projection dimension  & 192 \\
\midrule
GRN (dynamic)     & Pre-enrich.\ hidden   & 256 \\
                  & Post-enrich.\ hidden  & 128 \\
                  & Dropout (pre/post)    & 0.1\,/\,0.05 \\
                  & LayerNorm $\epsilon$  & $1{\times}10^{-5}$ \\
                  & Activation            & PReLU \\
                  & Gate mechanism        & Mult.\ \& additive \\
\midrule
GRU encoder       & Hidden state          & 256 \\
                  & Layers                & 2 \\
                  & Dropout               & 0.1 \\
                  & Activation            & tanh \\
\bottomrule
\end{tabular}
\end{minipage}
\hfill
\begin{minipage}[t]{0.49\textwidth}
\centering
\begin{tabular}{@{}l l l@{}}
\toprule
\textbf{Component} & \textbf{Parameter} & \textbf{Value} \\
\midrule
Multi-head attn.  & Number of heads       & 4 \\
                  & Head dim.\ $d_k$      & 32 \\
                  & Attention dropout     & 0.1 \\
                  & Normalisation         & LayerNorm \\
\midrule
Training          & Batch size            & 64 \\
                  & Initial LR            & $1{\times}10^{-3}$ \\
                  & LR decay factor       & 0.1 \\
                  & LR schedule           & ReduceLROnPlateau \\
                  & Early-stop patience   & 10 epochs \\
                  & Gradient clipping     & 1.0 \\
\midrule
Optimisation      & Optimiser             & Adam \\
                  & $\beta_1,\beta_2$     & 0.9,\,0.999 \\
\midrule
Regularisation    & Weight decay          & $1{\times}10^{-4}$ \\
                  & Dropout (overall)     & \shortstack[l]{0.1 pre-GRN/GRU/MHA;\\0.05 post-GRN} \\
\midrule
Model size        & Total parameters      & $\approx$11.4\,M \\
\bottomrule
\end{tabular}
\end{minipage}
\end{table*}

\section{Results and Discussion}
\label{sec:results}

\subsection{Long-Horizon Classification Performance}
Table~\ref{tab:comparison_multi} reports the completed 7-, 14-, and 30-day cross-model comparison. \TQRNNd{} leads all four fixed-threshold metrics at every horizon. At 7 days it reaches 76.09\% F1 compared with 72.20\% for the strongest F1 baseline, RoBERTa; at 14 days the margin widens to 76.76\% versus 70.49\%; and at 30 days \TQRNNd{} reaches 79.97\% F1, 80.18\% recall, 81.82\% precision, and 82.39\% accuracy, compared with RoBERTa at 77.50\% F1 and 77.89\% accuracy. The 30-day gain is therefore not produced by a precision--recall trade-off alone because both recall and precision improve.

\begin{figure}[t]
\centering
\begin{tikzpicture}
\begin{axis}[
  width=\columnwidth,
  height=5.8cm,
  xlabel={Forecast horizon (days)},
  ylabel={F1-score (\%)},
  xtick={0,1,2},
  xticklabels={7,14,30},
  xmin=-0.15,
  xmax=2.15,
  ymin=68,
  ymax=82,
  tick label style={font=\scriptsize},
  label style={font=\scriptsize},
  legend style={
    font=\scriptsize,
    at={(0.5,-0.28)},
    anchor=north,
    legend columns=2
  },
  grid=both,
  grid style={black!10},
  mark size=2pt,
  clip=false
]

\addplot[
  draw=none,
  fill=teal!18,
  forget plot
] coordinates {
  (0,72.20)
  (1,70.49)
  (2,77.50)
  (2,79.97)
  (1,76.76)
  (0,76.09)
} \closedcycle;

\addplot[
  orange!85!black,
  semithick,
  mark=square*
] coordinates {
  (0,72.20)
  (1,70.49)
  (2,77.50)
};
\addlegendentry{RoBERTa}

\addplot[
  blue!70!black,
  very thick,
  mark=*
] coordinates {
  (0,76.09)
  (1,76.76)
  (2,79.97)
};
\addlegendentry{\TQRNNd}

\node[
  font=\scriptsize,
  fill=white,
  inner sep=1pt
] at (axis cs:0,74.15) {$+3.89$};

\node[
  font=\scriptsize,
  fill=white,
  inner sep=1pt
] at (axis cs:1,73.63) {$+6.27$};

\node[
  font=\scriptsize,
  fill=white,
  inner sep=1pt
] at (axis cs:2,78.74) {$+2.47$};

\end{axis}
\end{tikzpicture}

\caption{F1-score comparison between \TQRNNd{} and the strongest baseline
across the 7-, 14-, and 30-day forecast horizons. The shaded region represents
the absolute F1-score difference between the two models. \TQRNNd{}
outperforms RoBERTa at every evaluated horizon, with the largest observed
difference of 6.27 percentage points at 14 days.}

\label{fig:perf}
\end{figure}
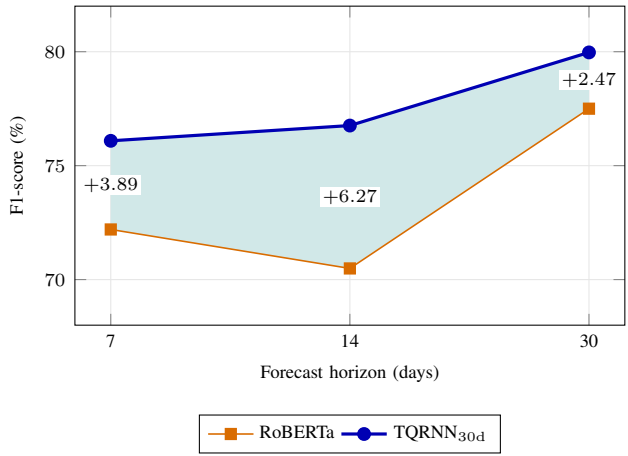

Figure~\ref{fig:perf} compares \TQRNNd{} with RoBERTa, the strongest
competing baseline at each evaluated horizon. \TQRNNd{} achieves F1-scores
of 76.09\%, 76.76\%, and 79.97\% at 7, 14, and 30 days, respectively, compared
with 72.20\%, 70.49\%, and 77.50\% for RoBERTa. The corresponding improvements
are 3.89, 6.27, and 2.47 percentage points. The advantage is therefore
maintained across all three horizons, although it is not monotonic and is
largest at 14 days.

\begin{table*}[t]
\centering
\caption{Complete long-horizon cross-model comparison at 7, 14, and 30 days under the common full-input protocol. The 18 baselines and proposed \TQRNNd{} are evaluated on the 15-machine held-out test cohort drawn from the 72-machine, 9-facility deployment. Values are percentages. Rows are ordered by mean F1 across the three horizons; bold denotes the strongest result at each horizon.}
\label{tab:comparison_multi}
\scriptsize
\setlength{\tabcolsep}{3.2pt}
\renewcommand{\arraystretch}{1.06}
\resizebox{\textwidth}{!}{%
\begin{tabular}{@{}lcccccccccccc@{}}
\toprule
& \multicolumn{4}{c}{\textbf{7-day}} & \multicolumn{4}{c}{\textbf{14-day}} & \multicolumn{4}{c}{\textbf{30-day}} \\
\cmidrule(lr){2-5}\cmidrule(lr){6-9}\cmidrule(lr){10-13}
\textbf{Model} & \textbf{F1} & \textbf{Rec.} & \textbf{Prec.} & \textbf{Acc.} & \textbf{F1} & \textbf{Rec.} & \textbf{Prec.} & \textbf{Acc.} & \textbf{F1} & \textbf{Rec.} & \textbf{Prec.} & \textbf{Acc.} \\
\midrule
MLP            & 52.60 & 52.31 & 54.80 & 53.91 & 46.79 & 48.17 & 49.59 & 50.27 & 53.75 & 54.21 & 54.44 & 54.81 \\
GAN            & 54.95 & 54.60 & 55.91 & 55.09 & 47.88 & 47.32 & 48.73 & 49.26 & 52.90 & 53.41 & 53.57 & 53.20 \\
Liquid         & 57.81 & 55.63 & 59.50 & 61.21 & 49.93 & 49.50 & 49.76 & 48.88 & 54.51 & 53.16 & 51.81 & 50.46 \\
LSRN           & 57.97 & 55.69 & 59.71 & 61.84 & 54.56 & 55.28 & 55.57 & 54.83 & 52.68 & 52.95 & 53.32 & 53.39 \\
KNN            & 62.52 & 62.18 & 62.67 & 63.54 & 54.62 & 55.35 & 55.61 & 55.03 & 53.42 & 53.82 & 54.10 & 54.00 \\
XGBoost        & 67.64 & 66.15 & 69.20 & 68.62 & 53.19 & 52.70 & 50.54 & 50.88 & 52.82 & 50.81 & 49.74 & 50.37 \\
LightGBM       & 65.30 & 64.45 & 65.71 & 65.80 & 55.89 & 56.71 & 57.08 & 56.39 & 54.92 & 55.32 & 55.45 & 55.17 \\
SVR            & 66.05 & 63.15 & 64.78 & 65.63 & 55.80 & 56.56 & 56.80 & 56.36 & 54.49 & 55.00 & 55.29 & 55.27 \\
LSTM           & 70.82 & 70.56 & 72.80 & 74.39 & 57.81 & 57.15 & 57.22 & 57.43 & 58.39 & 58.04 & 58.15 & 57.62 \\
LS-SVM         & 71.85 & 70.98 & 72.29 & 73.05 & 57.47 & 57.64 & 56.72 & 57.13 & 58.16 & 57.70 & 57.81 & 57.46 \\
Transformer    & 70.99 & 70.51 & 71.75 & 74.31 & 59.04 & 58.55 & 57.14 & 58.27 & 61.51 & 60.60 & 60.54 & 61.86 \\
DistilBERT     & 62.84 & 63.03 & 63.52 & 65.22 & 63.34 & 62.16 & 63.11 & 62.74 & 69.83 & 68.63 & 67.65 & 69.00 \\
TFT$_{\mathrm{full}}$ & 64.46 & 65.02 & 65.75 & 67.47 & 64.50 & 64.97 & 65.23 & 65.45 & 71.85 & 70.53 & 70.70 & 70.12 \\
Transformer-XL & 67.78 & 66.92 & 68.03 & 68.77 & 65.21 & 66.11 & 65.56 & 65.61 & 69.27 & 68.64 & 70.12 & 71.38 \\
BERT           & 66.05 & 65.73 & 66.61 & 65.24 & 65.28 & 65.47 & 66.00 & 66.47 & 71.44 & 70.12 & 69.28 & 69.68 \\
Autoencoder    & 70.13 & 70.15 & 70.33 & 72.08 & 66.54 & 65.69 & 64.22 & 63.77 & 68.92 & 69.28 & 69.53 & 69.33 \\
ALBERT         & 71.78 & 71.99 & 72.57 & 73.31 & 66.17 & 66.95 & 66.17 & 66.24 & 70.14 & 71.34 & 69.73 & 70.88 \\
RoBERTa        & 72.20 & 72.80 & 73.42 & 72.53 & 70.49 & 72.23 & 71.76 & 72.26 & 77.50 & 76.85 & 78.57 & 77.89 \\
\textbf{\TQRNNd{}} & \textbf{76.09} & \textbf{75.81} & \textbf{76.89} & \textbf{77.14} & \textbf{76.76} & \textbf{78.73} & \textbf{78.31} & \textbf{78.69} & \textbf{79.97} & \textbf{80.18} & \textbf{81.82} & \textbf{82.39} \\
\bottomrule
\end{tabular}}
\vspace{2pt}
\begin{minipage}{0.985\textwidth}
\footnotesize
\emph{Protocol note:} All models use the same upstream QRNN-derived representation and the same regime-specific held-out population. TFT$_{\mathrm{full}}$ is the cross-model TFT benchmark supplied with that representation; TFT$_{\mathrm{flat}}$ in Table~\ref{tab:ablation} removes the MLP--QRNN hierarchy and is therefore a different input condition.
\end{minipage}
\end{table*}

\subsection{Component Ablation}
\label{sec:ablation}

The completed ablation evidence is reported at two levels. Table~\ref{tab:quantile_sens} traces the day-scale pipeline from the transferred 70-hour reference through isolated QRNN stages and a Transformer-only configuration to the fully integrated \TQRNNd{}. Table~\ref{tab:ablation} separately isolates TFT-side fusion components under flat-input conditions. These complementary experiments should not be read as a one-component-at-a-time ablation of the final complete model.

\subsubsection{Secondary input-side ablation}
Table~\ref{tab:ablation} reports a separate TFT-side experiment conducted under
flat-input conditions. Its first row, TFT$_{\text{flat}}$, is the TFT
architecture with the MLP--QRNN hierarchy removed; it is therefore not
comparable with the TFT$_{\text{full}}$ row of
Table~\ref{tab:comparison_multi}, which receives the full quantile-state
representation. It is reported because it isolates the incremental value of the TFT-side fusion
components in the absence of the quantile representation. Because the input
condition differs from the full quantile-state benchmark, the table supports
TFT-side progression under that flat-input condition rather than single-component
causal attribution for the final \TQRNNd{} pipeline.

Under this controlled flat-input setting, TFT$_{\mathrm{flat}}$ reaches 51.06\% F1
at the 30-day horizon. Adding environmental context, historical look-back,
known-ahead covariates, and full TFT-side fusion increases the 30-day F1-score
to 62.34\%.

\begin{table}[t]
\centering
\caption{Compact TFT-side component ablation under flat-input, no-QRNN
conditions over 7-, 14-, and 30-day horizons.}
\label{tab:ablation}
\scriptsize
\setlength{\tabcolsep}{2.4pt}
\renewcommand{\arraystretch}{1.05}
\begin{tabular}{@{}lcccc@{}}
\toprule
\textbf{Config.} & \textbf{F1} & \textbf{Rec.} & \textbf{Prec.} & \textbf{Acc.} \\
\midrule
\multicolumn{5}{@{}l}{\textbf{7-day horizon}} \\
TFT$_{\text{flat}}$       & 54.40 & 55.07 & 53.93 & 54.95 \\
TFT$_2$                   & 56.39 & 57.11 & 55.82 & 56.87 \\
TFT$_3$                   & 58.74 & 59.13 & 58.29 & 59.17 \\
TFT$_4$               & 61.86 & 62.53 & 61.39 & 62.41 \\
TFT$_{\mathrm{All}}$    & \textbf{69.82} & \textbf{70.02} & \textbf{70.11} & \textbf{69.14} \\
\midrule
\multicolumn{5}{@{}l}{\textbf{14-day horizon}} \\
TFT$_{\text{flat}}$       & 53.31 & 53.87 & 54.18 & 53.43 \\
TFT$_2$                   & 55.21 & 55.88 & 56.21 & 55.33 \\
TFT$_3$                   & 57.67 & 58.27 & 58.83 & 57.68 \\
TFT$_4$               & 60.77 & 61.33 & 61.64 & 60.89 \\
TFT$_{\mathrm{All}}$    & \textbf{63.73} & \textbf{62.38} & \textbf{63.47} & \textbf{62.93} \\
\midrule
\multicolumn{5}{@{}l}{\textbf{30-day horizon}} \\
TFT$_{\text{flat}}$       & 51.06 & 49.50 & 50.70 & 51.37 \\
TFT$_2$                   & 52.19 & 50.56 & 51.72 & 52.41 \\
TFT$_3$                   & 54.76 & 52.76 & 54.09 & 54.61 \\
TFT$_4$               & 57.52 & 55.96 & 57.16 & 57.83 \\
TFT$_{\mathrm{All}}$    & \textbf{62.34} & \textbf{61.75} & \textbf{61.68} & \textbf{62.01} \\
\bottomrule
\end{tabular}

\begin{minipage}{\columnwidth}
\vspace{1mm}
\tiny
\emph{Note:} Values are percentages; bold denotes the best result per metric and
horizon. QRNN$_1$, QRNN$_2$, MLP$_1$, and MLP$_2$ are excluded throughout.
TFT$_{\text{flat}}$ is the TFT$_{\text{full}}$ architecture of
Table~\ref{tab:comparison_multi} without the MLP--QRNN quantile-state
representation. This table is secondary to
Table~\ref{tab:quantile_sens}. TFT$_2$ adds environmental/contextual
inputs; TFT$_3$ adds historical look-back; TFT$_4$ adds known-ahead covariates;
TFT$_{\mathrm{All}}$ combines TFT-side components only and is not the full
\TQRNNd{} pipeline.
\end{minipage}
\end{table}

\subsection{Threshold-Independent and Positive-Class Discrimination}

The threshold-swept results support the fixed-threshold findings.
Table~\ref{tab:roc_ref} shows the receiver-operating-characteristic curves at the
30-day horizon for the proposed model and the strongest comparators. At this
horizon \TQRNNd{} achieves the highest ROC--AUC, 0.820, compared with 0.724 for
TFT$_{\text{full}}$, 0.722 for RoBERTa, and 0.717 for Transformer-XL. The
separation is visible across the whole curve rather than at a single operating
point, which indicates improved class separation over the decision-threshold
range rather than a more favourable threshold choice. The 30-day AUC estimates were accompanied by prediction-instance
percentile-bootstrap 95\% intervals based on $B=10{,}000$ resamples:
$[0.819,0.821]$ for \TQRNNd{} and $[0.722,0.725]$ for
TFT$_{\mathrm{full}}$. Because consecutive documents can overlap temporally,
these intervals quantify prediction-population precision and are not interpreted
as machine-level, site-level, or unseen-domain generalisation intervals.

Positive-class discrimination is summarised using the mean multi-site
precision--recall profile area. As shown in Table~\ref{tab:pr_ref}, \TQRNNd{} achieves
the largest PR-profile area of 0.638,
ahead of RoBERTa (0.581) and Transformer-XL (0.552). Across the 19-model comparison (18 baselines plus \TQRNNd{}), the
corresponding all-model mean profile area was 0.432.

\begin{table}[t]
\centering
\caption{30-day ROC--AUC reference values.}
\label{tab:roc_ref}
\small
\begin{tabular}{@{}lc@{}}
\toprule
\textbf{Model} & \textbf{ROC--AUC}\\
\midrule
\TQRNNd{} & \textbf{0.820}\\
TFT$_{\mathrm{full}}$ & 0.724\\
RoBERTa & 0.722\\
Transformer-XL & 0.717\\
\bottomrule
\end{tabular}
\end{table}

\begin{table}[t]
\centering
\caption{Mean multi-site precision--recall profile areas at the 30-day horizon for selected strongest and weakest models from the 19-model comparison.}
\label{tab:pr_ref}
\small
\begin{tabular}{@{}lc@{}}
\toprule
\textbf{Model} & \textbf{PR-profile area}\\
\midrule
\TQRNNd{} & \textbf{0.638}\\
RoBERTa & 0.581\\
Transformer-XL & 0.552\\
KNN & 0.209\\
LightGBM & 0.201\\
Liquid & 0.140\\
\midrule
19-model mean & 0.432\\
\bottomrule
\end{tabular}
\end{table}

\subsection{Computational Cost}

Table~\ref{tab:compute} reports indicative computational cost so that the
accuracy gains can be weighed against deployment feasibility. Four quantities are
listed: training time per epoch (development cost), mean inference latency per
window (the cost that matters at deployment), parameter count (model capacity and
memory footprint), and peak GPU memory (the hardware required to serve the
model). Rows are grouped by model family and ordered by parameter count within
each group: the two proposed configurations first, then the transformer and
recurrent deep baselines, then the large pretrained language-model baselines, and
finally the lightweight classical methods. The comparison shows that although
\TQRNNd{} is heavier than compact classical and recurrent baselines, it remains
an order of magnitude smaller than the language-model baselines (11.4M parameters
versus 110--125M for BERT and RoBERTa) while achieving higher accuracy. Its
18~ms inference latency is well within the budget for batch-mode maintenance
planning, where predictions are refreshed at operational rather than millisecond
control timescales.

\subsubsection{Computational trade-off}
\label{sec:complexity}
\TQRNNd{} is heavier than several classical and recurrent baselines, so its
predictive gain should be interpreted alongside deployment cost. The comparison is
not simply a capacity effect: BERT and RoBERTa use roughly an order of magnitude
more parameters than \TQRNNd{} yet achieve lower 30-day classification scores.
The completed pipeline and TFT-side ablations provide complementary evidence that
the strongest result depends on both the QRNN-derived representation and the
subsequent temporal-fusion pathway, although they are not a matched one-component
complexity study. Finally, each document spans 30 days but a new document is
instantiated hourly. The serving requirement is therefore operational rather than a
millisecond control-loop constraint, and the principal computational burden lies in
model training and upstream data handling rather than the approximately 18~ms
classifier inference reported below.

\begin{table}[t]
\centering
\caption{Indicative computational cost, grouped by model family and ordered by
parameter count within each group. Lower is better for every column; values are
approximate and measured under a common evaluation setup.}
\label{tab:compute}
\small
\setlength{\tabcolsep}{4pt}
\begin{tabular}{lcccc}
\toprule
\textbf{Model} & \textbf{Train} & \textbf{Inf.} & \textbf{Params} & \textbf{GPU} \\
 & \textbf{(s/ep)} & \textbf{(ms)} & & \textbf{(GB)}\\
\midrule
\TQRNNd{}   & $\sim$310 & $\sim$18 & $\sim$11.4M & $\sim$14\\
\TQRNNh{}   & $\sim$85 & $\sim$8 & $\sim$3.2M  & $\sim$4\\
Transformer & $\sim$20 & $\sim$3 & $\sim$1.1M  & $\sim$2\\
LSTM        & $\sim$18 & $\sim$2 & $\sim$0.6M  & $\sim$1\\
RoBERTa     & $\sim$140 & $\sim$12 & $\sim$125M  & $\sim$18\\
BERT        & $\sim$130 & $\sim$11 & $\sim$110M  & $\sim$16\\
XGBoost     & $\sim$4 & $<$1     & N/A         & $<$1\\
SVR         & $\sim$6 & $<$1     & N/A         & $<$1\\
\bottomrule
\end{tabular}
\end{table}

\subsection{Operational Deployment Context}
\label{sec:deployment}

The framework was deployed within a broader maintenance and operational-
improvement programme across the nine participating facilities. During the
24-month observation period, the organisation's recorded production-output
efficiency increased from 78.38\% to 90.23\%. Internal accounting also indicated
reductions in maintenance disruption and scrap-related losses, together with
additional production output. Detailed financial values are omitted because
they depend on commercially sensitive cost, pricing, and production assumptions.

These observations provide deployment context rather than causal evidence of
model effectiveness. No contemporaneous untreated control facilities were
available, and changes to maintenance scheduling, operator practices, and
production processes occurred during the same period. The observed operational
and financial changes therefore cannot be attributed independently to the
predictive framework.

\subsection{Scope and Limitations}
\label{sec:limits}

The results should be interpreted within the following experimental and
deployment boundaries.

\paragraph{Fleet and transfer scope}
The evaluation covers 72 machines across nine facilities, but all machines
belong to the same asset and product family and use a common sensor
configuration. The study therefore assesses held-out machine performance within a homogeneous,
multi-facility industrial fleet; it does not establish transfer to an unseen
facility, different machine type, sensing architecture, product, or industrial
sector.

\paragraph{Operational horizons}
The 7-, 14-, and 30-day horizons were selected to represent progressively longer
maintenance, production-scheduling, and procurement planning windows. Although the experiments show that the
quantile-derived representation remains effective as the forecast interval
increases, the study does not derive a formal relationship between horizon
length, degradation observability, class separation, and attainable prediction
error. The reported horizon dependence should therefore be interpreted as an
empirical result supported by the mechanism discussed in
Section~\ref{sec:whyquantiles}, rather than as a general theoretical law.

\paragraph{Instability-aware gating}
The instability-aware pathway identifies sustained local divergence in the
prediction-error trajectory, but does not determine its physical cause. It may
therefore respond to sensor drift, material transitions, or sustained operating
regime changes as well as to progressive degradation. Its threshold and
persistence settings were calibrated using the training and validation data
from this deployment and would require recalibration before application to a
different machine population.

\paragraph{Dataset availability and temporal coverage}
The dataset was obtained from a proprietary industrial deployment and cannot be
released in its raw form, limiting independent reproduction. Although it spans
nine facilities and 24 months of operation from January 2024 to December 2025,
it remains a single organisational deployment. The evaluation consequently
does not establish performance under substantially different industrial
conditions or over longer periods of model ageing and concept drift.

\paragraph{Sensor selection}
The 81-channel configuration was determined from sensor availability,
engineering relevance, and prior operational knowledge rather than through a
formal feature-selection procedure. The sensor-ablation analysis evaluates the
sensitivity of the trained pipeline to individual and grouped channel loss, but
does not establish that the selected configuration is globally optimal.
Data-driven sensor-selection methods, such as \cite{ParallelGraphNet}, could
reduce sensing and computational requirements in future deployments.

\section{Conclusion}
\label{sec:conclusion}
This paper addressed long-horizon fault-occurrence forecasting in multi-site
industrial manufacturing and evaluated whether a conditional-quantile
representation provides a useful classifier interface across 7-, 14-, and 30-day
planning horizons.

\TQRNNd{} combines dual-stage QRNN feature extraction with a multi-stream temporal-
fusion classifier operating on hourly words and 30-day documents. The classifier
integrates quantile-state features, dynamic covariates, channel-level static
metadata, and latent history through gated processing, causal recurrent encoding,
and metadata-conditioned cross-modal attention. It produces structured hourly
forecast-interval probabilities from which the horizon-level Normal/Abnormal
forecast is derived. The auxiliary instability-aware pathway provides bounded
memory modulation during sustained local prediction-error divergence; it is a
supporting heuristic rather than a separately validated fault detector.

Under the 43/14/15 machine-disjoint train/validation/test allocation
across the 72-machine, nine-facility deployment, \TQRNNd{} attained 79.97\% F1,
80.18\% recall, 81.82\% precision, 82.39\% accuracy, and 0.820 ROC--AUC at 30
days. It led the completed 18-baseline fixed-threshold comparison at 7, 14, and
30 days. Relative to RoBERTa, the F1 improvements were 3.89, 6.27, and 2.47
percentage points, respectively. The advantage is therefore maintained across the
three evaluated horizons but is not monotonic; it is largest at 14 days.

The evidence supports long-horizon predictive maintenance within the evaluated
homogeneous machine family and observed nine-facility population. It does not
establish unseen-site, cross-equipment, or cross-sector generalisation. Likewise,
the operational improvements observed during deployment are contextual rather
than causal because no contemporaneous untreated facility group was available.
Future work should therefore prioritise genuinely facility-held-out validation,
a second machine family or public benchmark, multi-seed robustness analysis, and
prospective evaluation of maintenance decisions driven by the structured risk
trajectory.

\section*{Acknowledgment}
The authors thank the partner manufacturing organisation for access to deployment
data, operational context, and engineering support.

\section*{Data Availability}
The operational sensor and maintenance data used in this study are proprietary
and contain commercially sensitive information. The data cannot be made publicly
available under the data-sharing agreement with the industrial partner.
Aggregated results and sufficient methodological detail are reported to support
assessment of the experimental design and findings.




\begin{thebibliography}{00}

\bibitem{Sellitto}M.~A.~Sellitto and B.~Pinho, ``Maintenance strategy choice supported by the failure rate function: application in a serial manufacturing line,'' \emph{Period. Polytech. Soc. Manag. Sci.}, vol.~31, no.~1, pp.~38--51, 2023.

\bibitem{Sharma}R.~Sharma and B.~Vill\'anyi, ``Evaluation of corporate requirements for smart manufacturing systems using predictive analytics,'' \emph{Internet Things}, vol.~19, art.~100554, 2022.

\bibitem{Chen}C.~Chen, Z.~H.~Zhu, J.~Shi, N.~Lu, and B.~Jiang, ``Dynamic predictive maintenance scheduling using deep learning ensemble for system health prognostics,'' \emph{IEEE Sensors J.}, vol.~21, no.~23, pp.~26878--26891, 2021.

\bibitem{Abidi}M.~H.~Abidi, M.~K.~Mohammed, and H.~Alkhalefah, ``Predictive maintenance planning for Industry 4.0 using machine learning for sustainable manufacturing,'' \emph{Sustainability}, vol.~14, no.~6, art.~3387, 2022.

\bibitem{Yazdi}M.~Yazdi, ``Maintenance strategies and optimisation  techniques,'' in \emph{Advances in Computational Mathematics for Industrial System Reliability and Maintainability}. Cham: Springer, 2024, pp.~43--58.

\bibitem{Qiu}S.~Qiu, X.~Cui, Z.~Ping, N.~Shan, Z.~Li, X.~Bao, and X.~Xu, ``Deep learning techniques in intelligent fault diagnosis and prognosis for industrial systems: a review,'' \emph{Sensors}, vol.~23, no.~3, art.~1305, 2023.

\bibitem{Islam}M.~R.~Islam, S.~Begum, and M.~U.~Ahmed, ``Artificial intelligence in predictive maintenance: a systematic literature review on review papers,'' in \emph{Int. Congr. Workshop Ind. AI eMaintenance 2023}, Lecture Notes in Mechanical Engineering. Cham: Springer, 2024.

\bibitem{poland}
D.~J. Poland,
``Boosted Enhanced Quantile Regression Neural Networks with Spatiotemporal
Permutation Entropy for Complex System Prognostics,''
\emph{arXiv preprint arXiv:2507.14194}, 2025.

\bibitem{Serradilla}O.~Serradilla, E.~Zugasti, J.~Rodriguez, and U.~Zurutuza, ``Deep learning models for predictive maintenance: a survey, comparison, challenges and prospects,'' \emph{Appl. Intell.}, vol.~52, no.~10, pp.~10934--10964, 2022.

\bibitem{Pozzi}R.~Pozzi, T.~Rossi, and R.~Secchi, ``Industry 4.0 technologies: critical success factors for implementation and improvements in manufacturing companies,'' \emph{Prod. Plan. Control}, vol.~34, no.~2, pp.~139--158, 2023.

\bibitem{Fernandes}M.~Fernandes, J.~M.~Corchado, and G.~Marreiros, ``Machine learning techniques applied to mechanical fault diagnosis and fault prognosis in real industrial manufacturing use-cases: a systematic literature review,'' \emph{Appl. Intell.}, vol.~52, no.~12, pp.~14246--14280, 2022.

\bibitem{Mitici}M.~Mitici, I.~de~Pater, A.~Barros, and Z.~Zeng, ``Dynamic predictive maintenance for multiple components using data-driven probabilistic RUL prognostics: the case of turbofan engines,'' \emph{Reliab. Eng. Syst. Saf.}, vol.~234, art.~109199, 2023.

\bibitem{Meddaoui}A.~Meddaoui, M.~Hain, and A.~Hachmoud, ``The benefits of predictive maintenance in manufacturing excellence,'' \emph{Int. J. Adv. Manuf. Technol.}, vol.~128, no.~7--8, pp.~3685--3690, 2023.

\bibitem{Ong}K.~S.~Ong, W.~Wang, N.~Q.~Hieu, D.~Niyato, and T.~Friedrichs, ``Predictive maintenance model for IIoT-based manufacturing: a transferable deep reinforcement learning approach,'' \emph{IEEE Internet Things J.}, vol.~9, no.~17, pp.~15725--15741, 2022.

\bibitem{Mohammad}M.~S.~Jahangir, J.~You, and J.~Quilty, ``A quantile-based encoder--decoder framework for multi-step ahead runoff forecasting,'' \emph{J. Hydrol.}, vol.~619, 2023.

\bibitem{Grigsby}J.~Grigsby, Z.~Wang, N.~Nguyen, and Y.~Qi, ``Long-range transformers for dynamic spatiotemporal forecasting,'' \emph{arXiv:2109.12218}, 2021.

\bibitem{Vaswani}A.~Vaswani \emph{et al.}, ``Attention is all you need,'' in \emph{Proc. NeurIPS}, 2017.

\bibitem{Lim2021}B.~Lim, S.~\"O.~Ar\i k, N.~Loeff, and T.~Pfister, ``Temporal fusion transformers for interpretable multi-horizon time series forecasting,'' \emph{Int. J. Forecast.}, vol.~37, no.~4, pp.~1748--1764, 2021.


\bibitem{Wu}J.~Wu, Z.~Zhao, C.~Sun, R.~Yan, and X.~Chen, ``Fault-attention generative probabilistic adversarial autoencoder for machine anomaly detection,'' \emph{IEEE Trans. Ind. Informat.}, vol.~16, no.~12, pp.~7479--7488, 2020.

\bibitem{Oliosi}E.~Oliosi, G.~Calzavara, and G.~Ferrari, ``On sensor data clustering for machine status monitoring and its application to predictive maintenance,'' \emph{IEEE Sensors J.}, vol.~23, no.~9, pp.~9620--9639, 2023.

\bibitem{Carrasco}J.~Carrasco \emph{et al.}, ``Anomaly detection in predictive maintenance: a new evaluation framework for temporal unsupervised anomaly detection algorithms,'' \emph{arXiv:2105.12818}, 2021.

\bibitem{Iqbal}A.~Iqbal and R.~Amin, ``Time series forecasting and anomaly detection using deep learning,'' \emph{Comput. Chem. Eng.}, vol.~182, art.~108560, 2024.

\bibitem{Shoorkand}H.~D.~Shoorkand, M.~Nourelfath, and A.~Hajji, ``A hybrid CNN--LSTM model for joint optimisation of production and imperfect predictive maintenance planning,'' \emph{Reliab. Eng. Syst. Saf.}, vol.~241, art.~109707, 2024.

\bibitem{Gawde}S.~Gawde, S.~Patil, S.~Kumar, P.~Kamat, and K.~Kotecha, ``An explainable predictive maintenance strategy for multi-fault diagnosis of rotating machines using multi-sensor data fusion,'' \emph{Decis. Anal. J.}, vol.~10, art.~100425, 2024.

\bibitem{Li1}Y.~Li, Z.~Tao, L.~Wang, B.~Du, J.~Guo, and S.~Pang, ``Digital twin-based job shop anomaly detection and dynamic scheduling,'' \emph{Robot. Comput.-Integr. Manuf.}, vol.~79, art.~102443, 2023.

\bibitem{De}A.~De~Benedictis, F.~Flammini, N.~Mazzocca, A.~Somma, and F.~Vitale, ``Digital twins for anomaly detection in the industrial Internet of Things,'' \emph{IEEE Trans. Ind. Informat.}, vol.~19, no.~12, pp.~11553--11563, 2023.

\bibitem{He}B.~He and K.~J.~Bai, ``Digital twin-based sustainable intelligent manufacturing: a review,'' \emph{Adv. Manuf.}, vol.~9, no.~1, pp.~1--21, 2021.

\bibitem{Wolf}A.~Wolf, J.~B.~Swift, H.~L.~Swinney, and J.~A.~Vastano, ``Determining Lyapunov exponents from a time series,'' \emph{Physica D}, vol.~16, no.~3, pp.~285--317, 1985.

\bibitem{ZSTT}J.~Gao, S.-P.~Wang, R.~Chen, C.~Zhang, E.~Zio, Y.~Zhang, and Y.~Lu, ``ZSTT: A zero-shot time-series prediction model based on transformer and its application to hydraulic actuator performance degradation prediction,'' \emph{IEEE/ASME Trans. Mechatron.}, vol.~31, no.~3, pp.~2946--2957, 2026, doi: 10.1109/TMECH.2025.3630496.

\bibitem{DualBranch}Y.~Zhuo, Y.~Luo, T.~Chen, Y.~Zhang, Z.~Zheng, and M.~Zhu, ``Remaining useful life prediction of complex equipment based on adaptive dual-branch multi-scale feature fusion model,'' \emph{Mech. Syst. Signal Process.}, vol.~254, art.~114389, 2026, doi: 10.1016/j.ymssp.2026.114389.

\bibitem{ShipRUL}Y.~Ren, L.~Zhang, and P.~Shi, ``Adaptive cooperative intra--inter domain adversarial network with stage division for RUL prediction of ship electric propulsion system,'' \emph{IEEE Trans. Ind. Electron.}, vol.~73, no.~8, pp.~12477--12489, 2026, doi: 10.1109/TIE.2026.3672810.

\bibitem{ParallelGraphNet}Z.~Wang, H.~Zhang, L.~Qiu, S.~Zhang, J.~Qian, F.~Xiang, Z.~Pan, and J.~Tan, ``Towards high-speed elevator fault diagnosis: A ParallelGraphNet-driven multi-sensor optimisation selection method,'' \emph{Mech. Syst. Signal Process.}, vol.~228, art.~112450, 2025, doi: 10.1016/j.ymssp.2025.112450.




\end{thebibliography}
\end{document}